\documentclass[letterpaper, 10 pt, conference]{ieeeconf}  

\IEEEoverridecommandlockouts                              

\usepackage{graphics} 
\usepackage{epsfig} 
\usepackage{subfigure}
\usepackage{amsmath} 
\usepackage{epstopdf}
\usepackage{lipsum,amsmath,multicol}
\usepackage{float}
\usepackage{algorithm}
\usepackage{algorithmicx}
\usepackage{algpseudocode}
\usepackage{tcolorbox}

\newtheorem{theorem}{Theorem}

\title{\LARGE \bf
Inducing Multi-Convexity in Path Constrained Trajectory Optimization for Mobile Manipulators 
\vspace{-1ex}
}

\author{Arun Kumar Singh$^{1}$,  Andrei Ahonen $^{2}$, Reza Ghabcheloo $^{2}$,  Andreas Muller$^{3}$
\thanks{The research was partly supported by Estonian Center for IT Excellence . $^{1}$ TUIT, University of Tartu. ${^2}$ Tampere University, Finland, ${^3}$ Johannes Kepler University, Austria}
\vspace{-1ex}
}

\begin{document}

\maketitle
\vspace{-1ex}
\thispagestyle{empty}
\pagestyle{empty}

\allowdisplaybreaks

\begin{abstract}

In this paper, we propose a novel trajectory optimization algorithm for mobile manipulators under end-effector path, collision avoidance and various kinematic constraints. Our key contribution lies in showing how this highly non-linear and non-convex problem can be solved as a sequence of convex unconstrained quadratic programs (QPs). This is achieved by reformulating the non-linear constraints that arise out of manipulator kinematics and its coupling with the mobile base in a multi-affine form. We then use techniques from Alternating Direction Method of Multipliers (ADMM) to  formulate and solve the trajectory optimization problem. The proposed ADMM has two similar non-convex steps. Importantly, a convex surrogate can be derived for each of them. We show how large parts of our optimization can be solved in parallel providing the possibility of exploiting multi-core CPUs/GPUs. We validate our trajectory optimization on different benchmark examples. Specifically, we highlight how it solves the cyclicity bottleneck and provides a holistic approach where diverse set of trajectories can be obtained by trading-off different aspects of manipulator and mobile base motion.


\end{abstract}


\section{Introduction}

\noindent {\textbf{Motivation and Contribution}}

\noindent Robotic painting and 3D printing \cite{cuong_3dprinting} are some applications where mobile base and the manipulator needs to be moved simultaneously to ensure that the end-effector traces a given trajectory. Furthermore, joint space limits and collision free region for the mobile base needs to be taken into account while generating smooth joint trajectories. In this paper, we formulate the coordination of the mobile base and the manipulator as a trajectory optimization problem. Our key contribution lies in showing how this difficult non-linear non-convex optimization can be solved efficiently exploiting the underlying mathematical structures of the kinematics of the manipulator, the mobile base, and their coupling.

The proposed optimizer builds upon our recent work \cite{aks_iros18} and consists of two central ideas. The first among these involve introducing a few sets of auxiliary variables in which the manipulator kinematics and its non-linear coupling with the mobile base can be represented in a multi-affine form (see \cite{multiaffine_arxiv}, Section 2 for a light introduction to multi-affine/convex structure). Subsequently, techniques from Alternating Direction Method of Multipliers (ADMM) are used to formulate and solve the trajectory optimization problem. All but two steps of our ADMM involves solving an unconstrained convex quadratic programming (QP) problem (see Algorithm \ref{algo1}, \ref{algo2} ). The non-convex steps which have similar computational structures pertains to projecting the auxiliary variables back to the configuration space of the mobile manipulator. Consequently, our second key idea involves  deriving a class of convex surrogate to replace the non-convex steps of the proposed ADMM (see (\ref{theta_update}) and (\ref{phi_update})). The proposed trajectory optimization provides several key benefits over existing works:

\begin{itemize}
\item Compared to our prior work \cite{aks_iros18}, it has a significantly simpler derivation and enjoys better computational structure (see Section \ref{notes} ). Furthermore, it  also extends \cite{aks_iros18} to the case of mobile manipulators with either holonomic or non-holonomic base.

\item Closed cyclic trajectories in the configuration space of both the manipulator and the mobile base can be ensured corresponding to cyclic trajectories in the end-effector position space, extending our prior result on fixed base manipulator \cite{aks_iros18}.

\item In contrast to de-coupled approaches \cite{cuong_3dprinting}, we provide a holistic approach simultaneously optimizing different aspects of mobile base and manipulator motions using suitably defined cost functions.

\item We show that large parts of the proposed trajectory optimization have a distributive structure leading to the possibility of exploiting multi-core CPUs/GPUs.
\end{itemize}
%



\noindent \textbf{Related Works--Sampling Based Approach:} Constraints on end-effector path implicitly define a manifold in the configuration space and it is challenging to directly sample from this manifold. Nevertheless, existing works like  \cite{brenson_tcm}, \cite{rrt_manifold}, \cite{kuka_cyclic_sample} have successfully adapted sampling based planners like Rapidly Exploring Random trees (RRT) to account for task constraints. In  \cite{brenson_tcm}, authors use gradient descent based on manipulator Jacobian to project randomly sampled configurations to manifolds defined by the task constraints. In contrast, \cite{rrt_manifold} exploits locally valid affine parametrization of constraint manifolds to directly construct RRT on them. This approach was extended in \cite{kuka_cyclic_sample} to solve the cyclicity bottleneck in redundant manipulators, i.e, ensuring that closed cyclic trajectories in end-effector position space result in similar trajectories in the configuration space. Since these cited works are built on RRT, they do not incorporate any notion of optimality. Furthermore, trajectories computed needs to be post-processed to ensure smoothness \cite{brenson_tcm}. In contrast, the proposed trajectory optimization can directly compute trajectories with any desired level of differentiability. 

\noindent \textbf{Related Works--Optimization Based Approach:} In this approach, end-effector path constraints are modeled as non-linear equality constraints. Optimizers like CHOMP \cite{chomp}, \cite{chomp_modified}, TrajOpt \cite{trajopt} can handle such constraints and are also applicable to manipulators mounted on a holonomic base.  To the best of our knowledge, there are no applications of these cited trajectory optimizers to non-holonomic mobile manipulators. A recent work \cite{jonas_nonhol} is closely related to the current proposed work as it explicitly considers the non-holonomic constraints within the mobile manipulation. However, trajectory optimizer proposed in \cite{jonas_nonhol} does not consider inequality constraints due to joint limits and collision avoidance for the mobile base.  
Optimal control methods based on dynamic programming were proposed \cite{KasererTRO} that are able to take into account joint limits (vel., acc., etc.) for stationary robots.

\noindent \textbf{Related Works--Convexity in Manipulator Kinematics:} Recent works like \cite{convex_relaxation1}, \cite{convex_relaxation2} have made strong attempts towards  deriving convex relaxations for manipulator inverse kinematics problem. Our formulation has some key differences with both these cited works. First, we note that the solution of the relaxed convex problem may not be feasible with respect to the original non-convex problem. In contrast, we derives convex surrogates whose solutions exactly corresponds to the original problem. However, the convex surrogates may have access to a reduced search space (see Fig.\ref{convex surrogate}). Second, \cite{convex_relaxation1} is valid for only planar manipulators in general and some specific spatial manipulators. In contrast, our formulation makes no assumptions on the nature of the kinematic structure of the manipulators. Finally, \cite{convex_relaxation1}, \cite{convex_relaxation2} has not been extended to non-holonomic mobile manipulators.


\vspace{-0.2cm}

\section{Proposed Trajectory Optimization and ADMM}
\vspace{-0.1cm}

\begin{table}[!t]
\scriptsize
\centering
\caption{Important Symbols  } \label{sym_not}
\begin{tabular}{|p{2.1cm}|p{5cm}|p{5cm}|p{5cm}|}\hline
$\{0\}, \{l\}, \{g\} $ & Manipulator base, mobile base and global reference frame respectively.\\ \hline
${^l}\textbf{x}_o = ({^l}x_o, {^l}y_o, 0)$ & Position of the origin of $\{0\}$ with respect to the origin of $\{l\}$ expressed in the reference frame of the later.    \\ \hline
${^g}\textbf{x}_{b} = ({^g}x_{b}, {^g}y_{b}, 0)$ & Mobile base center in the global frame \\ \hline
${^l}\textbf{x}_{e} = ({^b}x_{e}, {^b}y_{e}, {^b}z_{e})$ & Vector from origin of $\{0\}$ to end-effector at time $t$ in the frame of the mobile  base\\\hline  
${^g}\textbf{x}_{e}$ & End-effector position in the global frame\\\hline
$\phi_{b}$ & Heading of the mobile base with respect to the global frame at time $t$\\ \hline
$\boldsymbol{\theta} = (\theta_{i}, \theta_{i} \dots \theta_{n})$ &  Vector of joint angles of the manipulator.\\ \hline
${^g}\textbf{x}_{d} = ({^g}x_{d}, {^g}y_{d}, {^g}z_{d})$ & Desired trajectory in the global frame for ${^g}\textbf{x}_{e}$\\\hline
\end{tabular}
\vspace{-0.8cm}
\normalsize
\end{table}

\noindent \textbf{Symbols and Notations:} We will use italic letters to represent scalars. Bold faced lower case letters will represent vectors while   upper case variants will represent matrices. Table \ref{sym_not} summarizes the important symbols used in the paper. Some symbols are also defined at their first place of use. We do not explicitly show the time dependency of the vectors, matrices and other variables.
We use a left superscript of $0$, $l$ and $g$ to denote whether a vector/matrix is defined in the manipulator base, mobile base or the global frame respectively. For notational simplicity and where it is obvious, we remove the subscripts defining the reference frame, e.g $\phi_{b}, \boldsymbol{\theta}$. The right superscript $T$ will represent transpose of a matrix or a row/column vector.   



\noindent \textbf{Assumptions:} (i): We assume that the mobile base operates in the $x-y$ plane. (iii): Collision avoidance is considered for the mobile base while the manipulator is assumed to be moving in the free space. This is typical of applications like robotic 3D printing or painting \cite{cuong_3dprinting}.

\noindent \textbf{Trajectory Optimization} The proposed trajectory optimization has the following form:

\vspace{-0.5cm}

\small
\begin{subequations}
\begin{align}
\arg\min w_1\overbrace{\sum_t \Vert \ddot{\boldsymbol{\theta}}\Vert_2^2}^{J_{man}} + w_2\overbrace{\sum_t \Vert \dot{\textbf{x}}_{b}\Vert_2^2}^{J_{base}} \label{cost}   \\
\textbf{f}_m(\boldsymbol{\theta}, {^g}\textbf{x}_{b},  {^g}\dot{\textbf{x}}_{b}, \phi_b) = \textbf{0}, \forall t,  \forall m \label{nonlin}\\
(\boldsymbol{\theta}, \dot{\boldsymbol{\theta}}, \ddot{\boldsymbol{\theta}}) \in \mathcal{C}_{\theta}, (\dot{\textbf{x}}_b, \ddot{\textbf{x}}_b ) \in \mathcal{C}_{\textbf{x}_b},  \textbf{A}_{coll}\textbf{x}_b \leq \textbf{B}_{coll}, \forall t \label{feasible_set}
\end{align}
\end{subequations}
\normalsize


\noindent The cost function (\ref{cost}) consists of two terms corresponding to the manipulator ($J_{man}$) and the base ($J_{base}$) motions. 
The weights $w_1, w_2$ trades-off contribution from each cost term. The cost $J_{man}$ is modeled as the sum of squared accelerations based on previous works like \cite{aks_iros18}, \cite{toussaint_NM} and acts as a  simpler surrogate for minimizing torques. The cost term ($J_{base}$) penalizes the norm of the velocities to limit the distance traveled by the mobile base. 

The constraints (\ref{nonlin}) represent a set of $m$ highly non-linear  equalities that models various constraints on the end-effector and the mobile base. We present a detailed analysis of these constraints in the next section.  The terms $\mathcal{C}_{\theta}$ and $\mathcal{C}_{\textbf{x}_b}$ represent the feasible set for $\boldsymbol{\theta}$, $\textbf{x}_b$ and their derivatives. These are defined by the affine equalities and inequalities modeling the boundary conditions and joint limits \footnote{We do not incorporate any bounds on velocities and accelerations as these can be satisfied through time scaling based post-processing \cite{cuong_topp} }. 

The inequality in (\ref{feasible_set}) models the collision free regions for the mobile base. Our collision avoidance model is based on representing the mobile base as a circular disk and obstacles as ellipses with axis aligned with the global $x-y$. Consequently, collision avoidance take the form of purely concave quadratic inequalities \cite{boyd_ccp2}. We compute affine approximations (\ref{feasible_set}) of these quadratic inequalities. As shown in \cite{boyd_ccp2}, an affine approximation of a concave quadratic inequality acts as its upper bound. In other words, satisfaction of the affine approximation guarantees satisfaction of the original quadratic inequality and consequently collision avoidance. The affine approximations can be  improved at each iteration of the optimization \cite{boyd_ccp2}.

\noindent \textbf{Trajectory Parametrization}
\noindent We require that the manipulator joints and mobile base trajectories be smooth. This can be achieved by ensuring that these trajectories are representable in the following form:

\vspace{-0.4cm}
\small
\begin{subequations}
\begin{align}
\theta_i = \textbf{p}\textbf{c}_{\theta_i}, {^g}\textbf{x}_{b} = \textbf{P} \textbf{c}_{\textbf{x}_b}, \textbf{c}_{\textbf{x}_b} = \begin{bmatrix}
\textbf{c}_{x_b}\\
\textbf{c}_{y_b}
\end{bmatrix}\label{smoothness_criteria_1} \\
\textbf{p} = \begin{bmatrix}
\psi_1(t) & \psi_2(t) & \dots \psi_m(t) 
\end{bmatrix}, \textbf{P} = \begin{bmatrix}
\textbf{p} & \textbf{0}\\
\textbf{0} & \textbf{p}
\end{bmatrix} \label{smoothness_criteria_2}
\end{align}
\end{subequations}
\normalsize

\noindent The vector $\textbf{p}$ is formed by smooth and differentiable time dependent basis functions $\psi(t)$ such as polynomials while the vectors $\textbf{c}_{\theta_i}$, $\textbf{c}_{x_b}$, $\textbf{c}_{y_b}$ are the coefficients associated with the basis functions.

There are two possible ways to incorporate the above parametrization within (\ref{cost})-(\ref{feasible_set}). We can directly replace $\theta_i$, ${^g}\textbf{x}_{b}$ and their derivatives by suitable functions of $\textbf{c}_{\theta_i}$ and $\textbf{c}_{\textbf{x}_b}$ respectively and subsequently directly optimize in the space of these coefficients. On the other hand, we can retain both $\theta_i$, ${^g}\textbf{x}_{b}$ and  $\textbf{c}_{\theta_i}, \textbf{c}_{\textbf{x}_b}$ as optimization variables and include (\ref{smoothness_criteria_1}) as additional equality constraints (similar to the the multiple shooting approach in optimal control).

In our work, we adopt a hybrid set-up combining the two approaches. We directly replace the mobile base variables  ${^g}\textbf{x}_{b}$ and its derivatives with the help of  $\textbf{c}_{\textbf{x}_b}$. In contrast for the manipulator part, we retain both $\theta_i$  along with $\textbf{c}_{\theta_i}$ and formulate different parts of the trajectory optimization with either of these variables. Our proposed hybrid set-up has two main motivations: (i) We found this technique to be numerically more stable than the two approaches described above. (ii) As shown later, it leads to a heavily distributive structure in the trajectory optimization. 

Under parametrization (\ref{smoothness_criteria_1}), optimization (\ref{cost})-(\ref{feasible_set}) is re-written in the following form:

\vspace{-3ex}
\small
\begin{subequations}
\begin{align}
 \min w_{1} \overbrace{\sum_{t,i}\Vert \ddot{\textbf{p}}\textbf{c}_{\theta_i}\Vert_2^2}^{J_{man}} + w_{2} \overbrace{\sum_{t,i}\Vert \dot{\textbf{p}}\textbf{c}_{\textbf{x}_b}\Vert_2^2}^{J_{base}} \label{cost_task} \\
\textbf{f}_m(\boldsymbol{\theta},  \textbf{c}_{\textbf{x}_b}, \phi_b) = \textbf{0}, \forall t,  \forall m  \label{nonlin_task}\\
\theta_i = \textbf{p}\textbf{c}_{\theta_i}, \forall i, \forall t \label{theta_polyeq} \\
(\boldsymbol{\theta}, \dot{\boldsymbol{\theta}}, \ddot{\boldsymbol{\theta}}) \in \mathcal{C}_{\theta}  \Rightarrow \textbf{G}_{\theta_i} \textbf{c}_{\theta_i} = \textbf{h}_{\theta_i},
\textbf{A}_{\theta_i} \textbf{c}_{\theta_i}\leq \textbf{b}_{\theta_i}  \label{joint_ineq_task},  \forall i \\
(\dot{\textbf{x}}_b, \ddot{\textbf{x}}_b ) \in \mathcal{C}_{\textbf{x}_b} \Rightarrow \textbf{G}_{\textbf{x}_{b}}\textbf{c}_{\textbf{x}_b} = \textbf{h}_{\textbf{x}_{b}}\label{base_ineq_task}  \\ 
\widetilde{\textbf{A}}_{coll} \textbf{c}_{\textbf{x}_b}-\textbf{b}_{coll} \leq 0,  \widetilde{\textbf{A}}_{coll} = \textbf{A}_{coll}\textbf{P}, \forall t \label{collavoid}
\end{align}
\end{subequations}
\normalsize
\vspace{-0.6cm}

\noindent where, $\textbf{G}_{\theta_i}$, $\textbf{A}_{\theta_i}$ and $\textbf{G}_{\textbf{x}_{b}}$ are known matrices constructed from $\textbf{p}, \dot{\textbf{p}}$ etc. The, vectors $\textbf{h}_{\theta_i}$, $\textbf{h}_{\textbf{x}_{b}}$ are formed with boundary values for $\boldsymbol{\theta}$, ${^g}\textbf{x}_b$ and their derivatives. The vector $\textbf{b}_{\theta_i}$ is formed with  the known values for joint limits.  The optimization variables now consists of $\boldsymbol{\theta}$, $\phi_b$ along with the coefficients $\textbf{c}_{\theta_i}$ and  $\textbf{c}_{\textbf{x}_b}$.

\noindent \textbf{ADMM Based Solution:} ADMM based approaches are most commonly used for equality constrained optimization problems, where each equality is replaced with a quadratic penalty augmented with a Lagrange multiplier. To handle inequality constraints from (\ref{joint_ineq_task}) and (\ref{collavoid}), we introduce respective non-negative slack variables $\textbf{s}_{\theta_i}$,$\textbf{s}_{coll}$. The final form of ADMM based reformulation of (\ref{cost_task})-(\ref{collavoid}) is shown in (\ref{admm_formulation}). Herein, the various $\lambda$ and $\rho$ are the Lagrange multipliers and the quadratic penalty parameters, respectively. 
\vspace{-2ex}

\small
\begin{align}
\arg\min \mathcal{L} (\textbf{c}_{\theta_i}, \boldsymbol{\theta}, \textbf{c}_{\textbf{x}_b}, \textbf{c}_{\phi} ) =  \arg\min w_1J_{man}(\textbf{c}_{\theta_i} )+ w_2 J_{base}(\textbf{c}_{\textbf{x}_b})\nonumber \\ +\sum_{t,m} \textbf{f}_m^T \boldsymbol{\lambda}_m +  \rho_m \Vert \textbf{f}_m\Vert_2^2 \nonumber \\ + \sum_{t,i} (\theta_i-\textbf{p}\textbf{c}_{\theta_i}) {\lambda}_{\textbf{c}_{\theta_i}}+  \rho_{\textbf{c}_{\theta_i}} (\theta_i-\textbf{p}\textbf{c}_{\theta_i})^2 \nonumber \\ 
+\sum_i(\textbf{G}_{\theta_i} \textbf{c}_{\theta_i} -\textbf{h}_{\theta_i})^T \boldsymbol{\lambda}_{\textbf{G}}^{\textbf{c}_{\theta_i}} + \rho_{\textbf{G}}^{\textbf{c}_{\theta_i}} \Vert \textbf{G} \textbf{c}_{\theta_i} -\textbf{h}_{\theta_i} \Vert_2^2   \nonumber \\
+\sum_i(\textbf{A}_{\theta_i} \textbf{c}_{\theta_i} + \textbf{s}_{\theta_i} -\textbf{b}_{\theta_i})^T\boldsymbol{\lambda}_{\textbf{A}}^{\textbf{c}_{\theta_i}}+\rho_{\textbf{A}}^{\textbf{c}_{\theta_i}} \Vert\textbf{A}_{\theta_i} \textbf{c}_{\theta_i} + \textbf{s}_{\theta_i} -\textbf{b}_{\theta_i}\Vert_2^2 \nonumber \\
+(\textbf{G}_{\textbf{x}_b} \textbf{c}_{\textbf{x}_b} -\textbf{h}_{\textbf{x}_b})^T \boldsymbol{\lambda}_{\textbf{G}}^{\textbf{c}_{\textbf{x}_b}} +\rho_{\textbf{G}}^{\textbf{c}_{\textbf{x}_b}} \Vert \textbf{G}_{\textbf{x}_b} \textbf{c}_{\textbf{x}_b} -\textbf{h}_{\textbf{x}_b} \Vert_2^2   \nonumber \\
+\sum_t (\widetilde{\textbf{A}}_{coll} \textbf{c}_{\textbf{x}_b}+\textbf{s}_{coll}-\textbf{b}_{coll})^T \boldsymbol{\lambda}_{coll} \nonumber \\ +\rho_{coll}\Vert \widetilde{\textbf{A}}_{coll} \textbf{c}_{\textbf{x}_b}+\textbf{s}_{coll}-\textbf{b}_{coll}\Vert_2^2 \label{admm_formulation} 
\end{align}
\normalsize

The solution iterates of (\ref{admm_formulation}) are summarized in (\ref{step_1})-(\ref{step_3}). Herein, $(.)^k$ denotes the value of the variables at iteration $k$. The various $\lambda$ and $\rho$ are updated based on residuals of the equality constraints \cite{boyd_admm}. The slack variables can be updated following the process described in \cite{admm_qp}.  Step (\ref{step_1}) is simple as it involves solving a convex QP. Thus, the core complexity stems from steps (\ref{step_2})-(\ref{step_3}) which involves optimization over non-convex $\textbf{f}_m$. In the next section, we show how these two highly non-linear and non-convex optimizations can be replaced with convex QPs. 
\small
\begin{subequations}
\begin{align}
(\textbf{c}_{\theta_i})^{k+1} = \arg\min \mathcal{L} ( (\boldsymbol{\theta})^k, (\dot{\boldsymbol{\theta}})^k, (\textbf{c}_{\textbf{x}_b})^k, (\phi_b)^k ) \label{step_1}  \\
(\boldsymbol{\theta})^{k+1} = \arg\min \mathcal{L} ( (\textbf{c}_{\theta_i})^{k+1}, (\textbf{c}_{\textbf{x}_b})^k, (\phi_b)^k ) \label{step_2}  \\
 (\textbf{c}_{\textbf{x}_b})^{k+1}, (\phi_b)^{k+1} = \arg\min \mathcal{L} ( (\boldsymbol{\theta})^{k+1}, (\dot{\boldsymbol{\theta}})^{k+1}, (\textbf{c}_{\theta_i})^{k+1}) \label{step_3}
\end{align}
\end{subequations}
\normalsize

\section{Main Results}
In this section, we present our main theoretical results which is decomposing (\ref{step_2})-(\ref{step_3}) to a sequence of convex QPs. We begin by describing the various building blocks.

\subsection{Convex Surrogate}
\noindent In this subsection, we derive a simple yet effective convex surrogate for a specific non-convex optimization problem that later forms the basis for simplifying (\ref{step_2})-(\ref{step_3}). The concepts presented here formalizes and builds upon the brief introduction provided in \cite{aks_iros18}. Consider the following two optimization problems and an associated Theorem
\vspace{-0.2cm}
\small
\begin{subequations}
\begin{align}
r = \arg\min_{r} f, f = (\cos(r)-v)^2+(\sin(r)-w)^2 \label{f_nonconvex} \\
r = \arg\min_{r} \widetilde{f}, \widetilde{f} = (r-\arctan2(\frac{w}{v}))^2 \label{f_convex}
\end{align}
\end{subequations}
\normalsize

\begin{figure}[!h]
  \centering
   \subfigure[]{
    \includegraphics[width= 3.00cm, height=3.2cm] {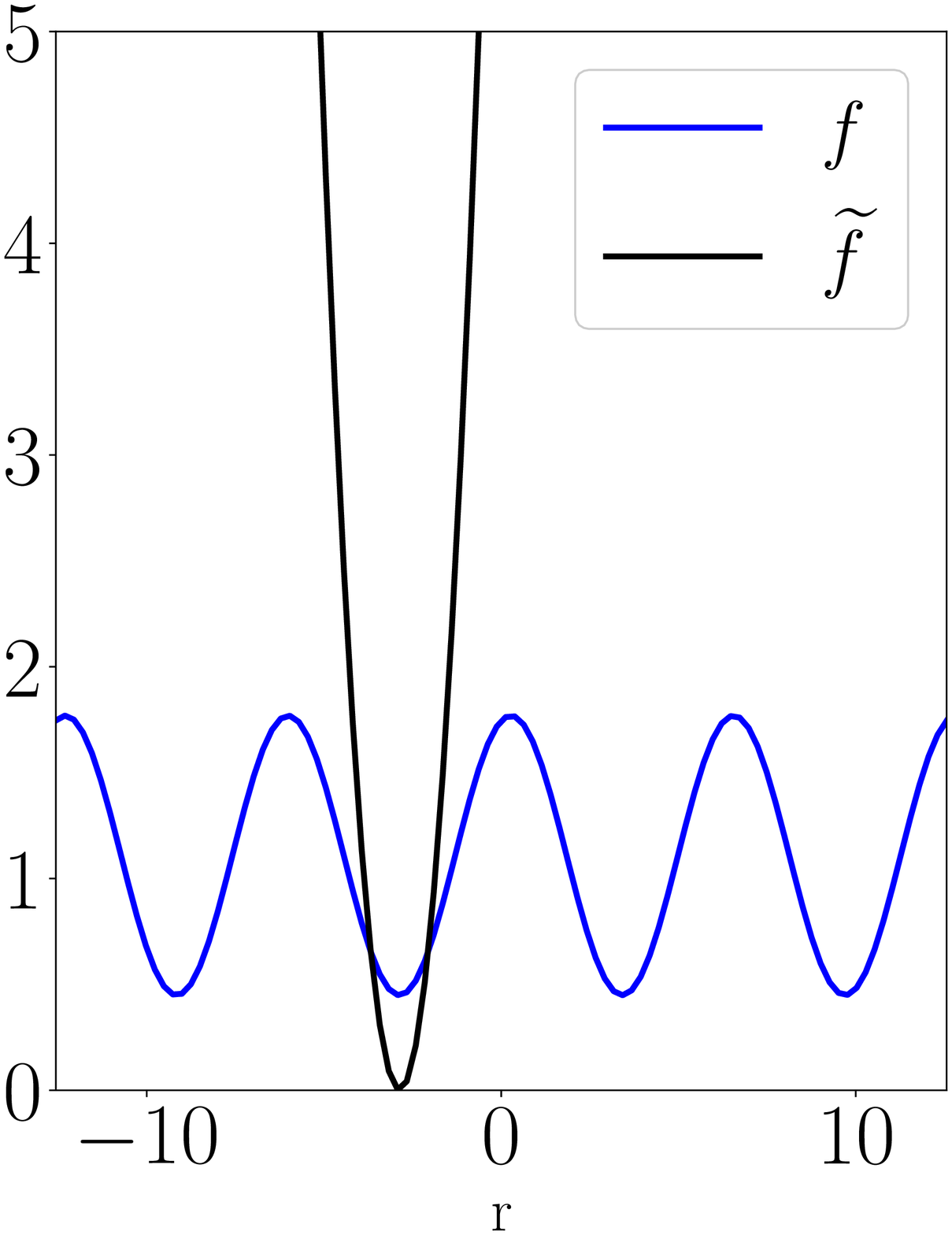}
    \label{convex surrogate}
   }\hspace{-0.60cm}
   \subfigure[]{
    \includegraphics[width= 5.00cm, height=3.2cm] {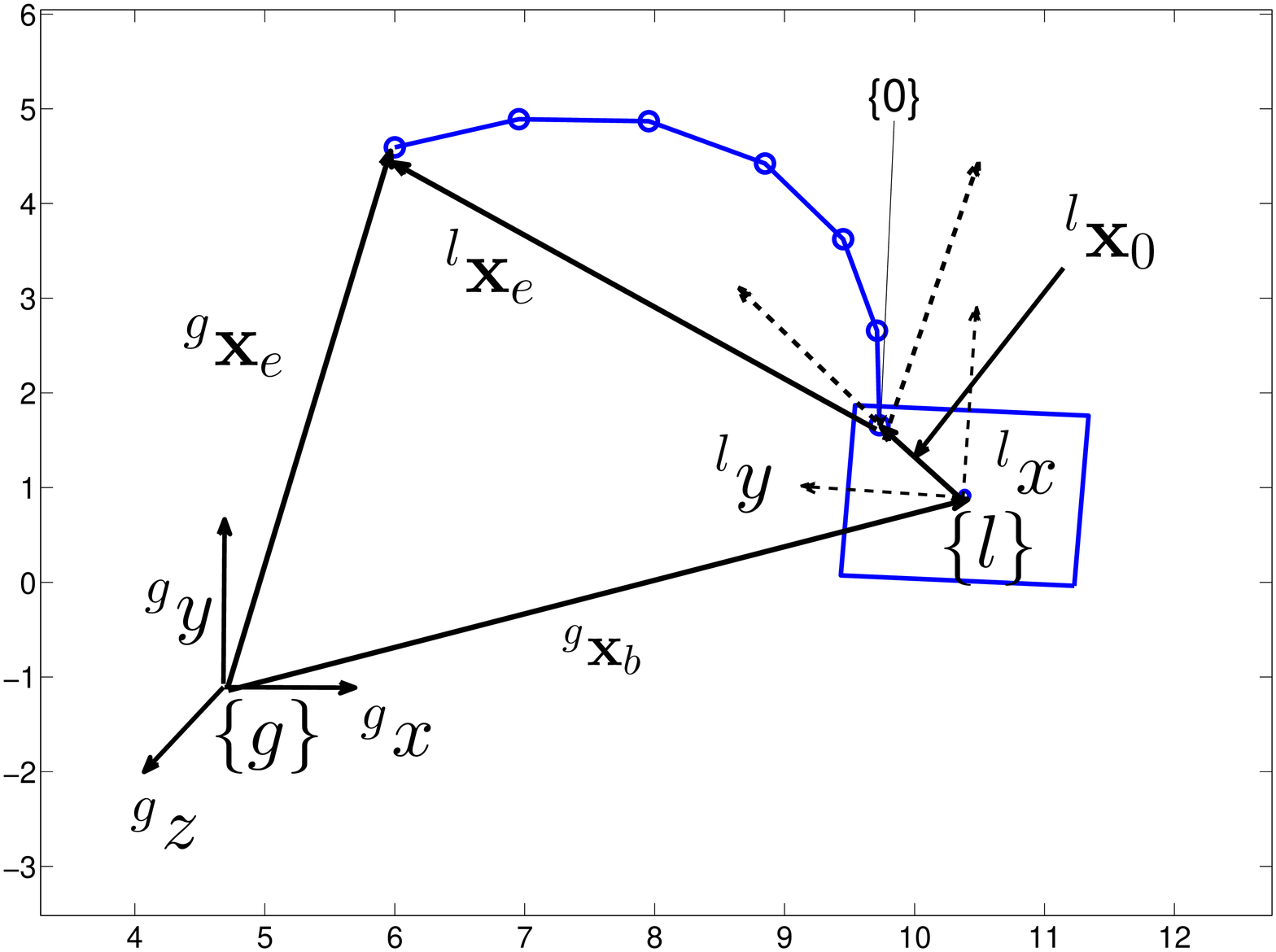}
    \label{mob_man_nom}
   }
\caption{(a): Plot of non-convex function $f$ and its convex surrogate $\widetilde{f}$. Minimizing $\widetilde{f}$ allows us to extract solution corresponding to one of the minima of $f$. (b): Relevant vectors for computing the end-effector position in the global frame }
\vspace{-0.5cm}
\end{figure}

\vspace{-0.2cm}
\noindent \begin{theorem}\label{convex_surrogate_theorem}
Optimization (\ref{f_nonconvex}) and (\ref{f_convex}) share a common minimizer (see Fig.\ref{convex surrogate}). 
\end{theorem}

\noindent \begin{proof}
Note that $\frac{df}{dr} = 0 \Rightarrow -u\sin(r) + v\cos(r) = 0$ and thus, $\frac{df}{dr}=0$ and $\frac{d\widetilde{f}}{dr}=0$ share a common solution and consequently a common minimizer. 
\end{proof}

An intuitive explanation of Theorem \ref{convex_surrogate_theorem} is presented in Fig.\ref{convex surrogate}. It can be seen that optimization (\ref{f_nonconvex}) has multiple local minima (albeit of equal optimal values). Optimization (\ref{f_convex}) acts as a convex surrogate that allows us to extract the solution corresponding to one of those minima. Now, the very nature of $\arctan2(.)$ suggests that the (\ref{f_convex}) will always extract the solution  that lies between   $[-\pi, \pi]$. However, this is not a strong limitation, if for example, $r$ represents rotation,  as $[-\pi, \pi]$ covers the full rotation range. For example, a rotation of say $6.2\pi$ can be a mapped to $0.2\pi$.  Furthermore, many industrial manipulators have joint motions within  $[-\pi, \pi]$.

\subsection{A Class of Non-Convex Optimization} 

\noindent Consider the following non-convex optimization problem:

\small
\begin{equation}
\min_{u,r}
f^2(u, \cos(r), \sin(r) )
\label{non-convex_ex}
\end{equation}
\normalsize

\noindent Assume that the function $f$ has a special structure: it is bi-affine in variables $u$ and the pair $(\cos(r), \sin(r))$. That is, for a fixed $u$, $f$ is affine simultaneously in $(\cos(r), \sin(r) )$. Similarly, fixing $r$  makes it affine in $u$. The  definition of $f$ can be extended to multi-affine case with arbitrary number of variables $u$ and $r$. For such a functional structure, a simple update rule of solving (\ref{non-convex_ex}) can be derived in the following manner. Introduce change of variables $v=\cos(r)$, $w = \sin(r)$ leading to  the following reformulation:

\vspace{-0.4cm}
\small
\begin{eqnarray}
\min f^2(u, v, w)
+\overbrace{\rho\Vert \begin{bmatrix}
\cos(r)\\
\sin(r)
\end{bmatrix}-\begin{bmatrix}
v\\
w
\end{bmatrix}\Vert_2^2}^{\text{consensus}}+\lambda_v v+\lambda_w w+\lambda_r r \label{reform_nonconvex_ex}
\end{eqnarray}
\normalsize
In (\ref{reform_nonconvex_ex}), we re-write $f$ in terms of $v, w$ and at the same time introduce a quadratic penalty which aims to bring a consensus between $v$ and $\cos(r)$ and $w$ and $\sin(r)$. The terms involving $\lambda_v, \lambda_w, \lambda_r$ are added to drive the consensus residuals to zero. The update rules for solving (\ref{reform_nonconvex_ex}) based on Gauss-Seidel (Alternating) minimization is given as
\vspace{-1ex}
\scriptsize
\begin{subequations} 
\begin{align}
(u)^{k+1} = \arg\min_u f^2((v)^k, (w)^k ) \label{step1_nonconvex} \\
(v, w)^{k+1} = \arg\min_{v,w}  f^2(u^{k+1}) 
+\rho\Vert \begin{bmatrix}
(\cos r)^k\\
(\sin r)^k
\end{bmatrix}- \begin{bmatrix}
v\\
w
\end{bmatrix}\Vert_2^2 
+\lambda_v v+\lambda_w w \label{step2_nonconvex}\\
(r)^{k+1} = \arg\min_r \Vert \begin{bmatrix}
(\cos(r))\\
(\sin(r))
\end{bmatrix}- \begin{bmatrix}
(v)^{k+1}\\
(w)^{k+1}
\end{bmatrix}\Vert_2^2+\lambda_r r \nonumber \\
= \rho\arg\min (r-\arctan2\frac{(w)^{k+1}}{(v)^{k+1}})^2+\lambda_r r. \label{step3_nonconvex} 
\end{align}
\end{subequations}
\normalsize
The first step (\ref{step1_nonconvex}) is convex and  minimizes $f^2(.)$ over $u$ while fixing $v, w$ at the values obtained in the previous iteration $k$. Step (\ref{step2_nonconvex}) is also convex and involves minimizing $f^2(.)$ over $v,w$ using $u$ obtained at the previous step. At the same time it also minimizes the consensus residuals. Step (\ref{step3_nonconvex}) is clearly non-convex but as shown, we can use Theorem \ref{convex_surrogate_theorem} to replace it with a convex surrogate. Step (\ref{step3_nonconvex}) can be seen as projecting $v, w$ back to the space of $r$.  
  
\noindent \textbf{Implication of (\ref{step1_nonconvex})-(\ref{step3_nonconvex}):} In the next subsection, we show that the various non-linear constraint functions $\textbf{f}_m$ in (\ref{nonlin_task}) can be reformulated to  have the same bi-affine or multi-affine structure as that of $f$ in (\ref{non-convex_ex}). Consequently, the optimization (\ref{step_2})-(\ref{step_3}) can be solved in the same manner as (\ref{step1_nonconvex})-(\ref{step3_nonconvex}).

\subsection{Multi-Affine Form}
\noindent In this subsection, we introduce each non-linear constraint function $\textbf{f}_m$ from (\ref{nonlin_task}) and if required, reformulate them to a multi-affine form. We begin by introducing the following change of variables from \cite{aks_iros18}, where cosine and sine of a vector translates to element wise cosine and sine.
\vspace{-0.2cm}
\small
\begin{subequations}
\begin{align}
v_{\theta_i} = \cos(\theta_i), \textbf{v}_{{\theta}} = (v_{\theta_1}, v_{\theta_2}, v_{\theta_n})= \cos(\boldsymbol{\theta}) \label{v_theta} \\
w_{\theta_i} = \sin(\theta_i), \textbf{w}_{{\theta}} = (w_{\theta_1}, w_{\theta_2}, w_{\theta_n})= \sin(\boldsymbol{\theta}) \label{w_theta}\\
v_{\phi} = \cos(\phi_{b}), w_{\phi} = \sin(\phi_{b}) \label{vw_phi} 
\end{align}
\end{subequations}
\normalsize

\noindent \textbf{$\textbf{f}_1$: } The function $\textbf{f}_1$ models the path constraints on the Cartesian position of end-effector. From Fig.\ref{mob_man_nom}, we derive the following loop-closure equation.

\vspace{-0.3cm}
\small
\begin{equation}
\textbf{f}_1 = {^g}\textbf{x}_{b}+{^g_l}\textbf{R}({^l}\textbf{x}_0+\overbrace{{^l_0}\textbf{R} \hspace{0.1cm}{^0}\textbf{x}_{e}}^{{^l}\textbf{x}_{e}  }) - {^g}\textbf{x}_{d} 
\label{x_task}
\end{equation}
\normalsize

\noindent Where, the rotation matrix ${^l_0}\textbf{R}$  depends on how the manipulator is connected to the mobile base and is constant. The term ${^g_l}\textbf{R}$ represents the rotation matrix between the local mobile base and the global frame. Now, using (\ref{smoothness_criteria_1}) and (\ref{vw_phi}), we can reformulate $\textbf{f}_1$ in the following form:

\vspace{-0.3cm}
\small
\begin{equation}
\textbf{f}_1 = \textbf{P}\textbf{c}_{\textbf{x}_b}+{\textbf{G}}_{\textbf{f}_1}({{^0}\textbf{x}_{e} }) \begin{bmatrix}
v_{\phi}\\
w_{\phi}\\
\end{bmatrix} -{^g}\textbf{x}_{d}
\label{x_task_reform2}
\end{equation}
\normalsize

\noindent Where, $\textbf{G}_{\textbf{f}_1}$ is a matrix whose elements are affine functions of ${^0}\textbf{x}_{e}$. Three sets of variables namely $\textbf{c}_{\textbf{x}_b}$, ${^0}\textbf{x}_e$, and $(v_{\phi}, w_{\phi} )$ can be easily identified from (\ref{x_task_reform2}). Fixing any two of these makes (\ref{x_task_reform2}) (affine) in the remaining set, thus highlighting the multi-affine structure.

\noindent {$\textbf{f}_2$:} In (\ref{x_task}), we treated ${^0}\textbf{x}_{e}$ as an independent variable. However, in actuality, it is exactly defined by the forward kinematics ($\textbf{f}_{fk}$) of the manipulator. We use constraint function $\textbf{f}_2$ to model this fact. 
\vspace{-0.25cm}
\small
\begin{equation}
\textbf{f}_2 =  {^0}\textbf{x}_{e} - \textbf{f}_{fk}(\boldsymbol{\theta})\label{man_fk}
\end{equation}
\normalsize

We now reformulate (\ref{man_fk}) in a bi-affine form.  The position forward kinematics of a $n$ degrees of freedom manipulator has the familiar form $ [{^0}\textbf{x}_{e}|1]^T = \prod {^{i-1}_i}\textbf{T}[\textbf{t}_c|1]^T $, where ${^{i-1}_i}\textbf{T}$ is the transformation matrix between joint $i$ and $i-1$ and $\textbf{t}_C$ is a vector defined in the end-effector reference frame.
Denote with ${^i}\textbf{x}_{e}$ the position vector of the end-effector measured from and resolved in the reference frame of joint $i$.
Then, (\ref{man_fk}) decomposes in the following form:
\small
\begin{subequations}
\begin{align}
{^{n-1}}\textbf{x}_{e} = {^{n-1}_n}\textbf{R}(c_{\theta_n}, s_{\theta_n})\textbf{t}_{c}+ {^{n-1}_n}\textbf{t} \label{xn-1} \\
{^{n-2}}\textbf{x}_{e} = {^{n-2}_{n-1}}\textbf{R}(c_{\theta_{n-1}}, s_{\theta_{n-1}}){^{n-1}}\textbf{x}_{e}+{^{n-2}_{n-1}}\textbf{t} \label{xn-1} \\
{^{n-3}}\textbf{x}_{e} = {^{n-3}_{n-2}}\textbf{R}(c_{\theta_{n-2}}, s_{\theta_{n-2}}){^{n-2}}\textbf{x}_{e}+{^{n-3}_{n-2}}\textbf{t} \label{xn-2}\\
\dots \dots \dots \dots \nonumber \\
{^1}\textbf{x}_{e} = {^1_2}\textbf{R}(c_{\theta_2}, s_{\theta_2}) {^2}\textbf{x}_{e}+ {^1_2}\textbf{t} \\
{^0}\textbf{x}_{e} = {^{0}_{1}}\textbf{R}(c_{\theta_1}, s_{\theta_1}){^1}\textbf{x}_{e}+{^{0}_{1}}\textbf{t}\label{x1}
\end{align}
\end{subequations}
\normalsize

\noindent Where ${^{i-1}_{i}}\textbf{R}(.)$, ${^{i-1}_{i}}\textbf{t}$ are respectively the rotation matrix and translation vector extracted from transformation matrix ${^{i-1}_{i}}\textbf{T}$ and $c_{\theta_{i}} = \cos(\theta_i), s_{\theta_i} = \sin(\theta_i) $. A slight algebraic manipulation using (\ref{v_theta})-(\ref{w_theta})  can put (\ref{xn-1})-(\ref{x1}) and consequently $\textbf{f}_2$ in the following form: 
\vspace{-0.2cm}
\small
\begin{equation}
\textbf{f}_2 = \textbf{G}_{\textbf{f}_2}({^i}\textbf{x}_{e} ) \begin{bmatrix}
\textbf{v}_{{\theta}}\\
\textbf{w}_{{\theta}}
\end{bmatrix}-\textbf{h}_{\textbf{f}_2}
\label{bi_convex_fk}
\end{equation}
\normalsize

\noindent Where $\textbf{G}_{\textbf{f}_2}$ is a matrix whose elements are affine functions of ${^i}\textbf{x}_e$. The vector $\textbf{h}_{\textbf{f}_2}$ is constant constructed from ${^{i-1}_{i}}\textbf{t}$ and $\textbf{t}_c$. Equation (\ref{bi_convex_fk}) clearly shows that $\textbf{f}_2$ is bi-affine with respect to two sets of variables ${^i}\textbf{x}_e$ and $(\textbf{v}_{{\theta}},\textbf{w}_{{\theta}})$.

\noindent {$\textbf{f}_3(.)$:} If the mobile base is non-holonomic, we use the final constraint function $\textbf{f}_3$ to model the no-lateral slip constraint i.e $\dot{x}_{b}\sin(\phi_{b})-\dot{y}_{b}\cos(\phi_{b}) =0$. We can write it in the following form using (\ref{vw_phi})

\vspace{-0.2cm}
\small
\begin{equation}
\textbf{f}_3 = \textbf{G}_{\textbf{f}_3} (\textbf{c}_{\textbf{x}_b}) \begin{bmatrix}
v_{\phi}\\
w_{\phi}\\
\end{bmatrix}
\label{nonhol_reform}
\end{equation}
\normalsize

\noindent Where, $\textbf{G}_{\textbf{f}_3}$ is a matrix whose elements are affine in $\dot{\textbf{x}}_{b}$ and consequently $\textbf{c}_{\textbf{x}_b}$. Clearly $\textbf{f}_3$ is bi-affine with respect to $\textbf{c}_{\textbf{x}_b}$ and $(v_{\phi}, w_{\phi})$.

\noindent \textbf{Summary:} From the discussions presented in this subsection, we can conclude that set of non-linear constraint functions  
have the form $\textbf{f}_m({^i}\textbf{x}_{e}, (\textbf{v}_{\boldsymbol{\theta}}, \textbf{w}_{\boldsymbol{\theta}}), \textbf{c}_{\textbf{x}_b}, (v_{\phi}, w_{\phi}) )$.


\noindent Importantly, $\textbf{f}_m$ is multi-affine in the variable sets ${^i}\textbf{x}_{e}$, $(v_{\phi}, w_{\phi})$, $(\textbf{v}_{\boldsymbol{\theta}}, \textbf{w}_{\boldsymbol{\theta}})$, $\textbf{c}_{\textbf{x}_b}$. Fixing any three of these variables makes $\textbf{f}_m$ affine in the remaining set.

\subsection{Simplifying (\ref{step_2})-(\ref{step_3})} 
\noindent The multi-affine structure coupled with reformulations presented in (\ref{reform_nonconvex_ex}) allows us to reformulate (\ref{step_2}) in the following form

\vspace{-0.6cm}
\small
\begin{align}
(\boldsymbol{\theta})^{k+1} = \arg\min \mathcal{L} ( (\textbf{c}_{\theta_i})^{k+1}, (\textbf{c}_{\textbf{x}_b})^k, (\phi_b)^k ) \nonumber \\
= \arg\min \sum_{t,m} \textbf{f}_m^T \boldsymbol{\lambda}_m +  \rho_m \Vert \textbf{f}_m\Vert_2^2\nonumber \\
+\sum_{t,i} (\theta_i-\textbf{p}(\textbf{c}_{\theta_i})^{k+1}) {\lambda}_{\textbf{c}_{\theta_i}}+  \rho_{\textbf{c}_{\theta_i}} (\theta_i-\textbf{p} (\textbf{c}_{\theta_i})^{k+1} )^2 \nonumber \\
+\sum_t \textbf{v}_{\theta}^T \boldsymbol{\lambda}_{\textbf{v}_\theta} + \textbf{w}_{{\theta}}^T \boldsymbol{\lambda}_{\textbf{w}_\theta}
+\rho_{\textbf{v}_\theta, \textbf{w}_\theta}\Vert \begin{bmatrix}
\cos(\boldsymbol{\theta})\\
\sin(\boldsymbol{\theta})\\
\end{bmatrix}- \begin{bmatrix}
\textbf{v}_{\boldsymbol{\theta}}\\
\textbf{w}_{\boldsymbol{\theta}}
\end{bmatrix}\Vert_2^2 + \boldsymbol{\theta}^T \boldsymbol{\lambda}_{\theta} \label{step_2_reform}
\end{align}
\normalsize

\noindent The first two lines in (\ref{step_2_reform}) are obtained by extracting $\boldsymbol{\theta}$ dependent terms from (\ref{admm_formulation}). Just like in (\ref{reform_nonconvex_ex}), the last line ensures consensus between $\textbf{v}_{\theta}$ and $\cos(\boldsymbol{\theta})$ and $\textbf{w}_{\theta}$ and $\sin(\boldsymbol{\theta})$.  A similar reasoning leads us to the following reformulation of optimization (\ref{step_3}).

\vspace{-0.3cm}
\small
\begin{align}
(\textbf{c}_{\textbf{x}_b})^{k+1}, (\phi_b)^{k+1} = \arg\min \mathcal{L} ( (\boldsymbol{\theta})^{k+1}, (\textbf{c}_{\theta_i})^{k+1}) \nonumber \\
\arg\min \sum_{t,m} \textbf{f}_m^T \boldsymbol{\lambda}_m +  \rho_m \Vert \textbf{f}_m\Vert_2^2 \nonumber \\
+(\textbf{G}_{\textbf{x}_b} \textbf{c}_{\textbf{x}_b} -\textbf{h}_{\textbf{x}_b})^T \boldsymbol{\lambda}_{\textbf{G}}^{\textbf{c}_{\textbf{x}_b}} +\rho_{\textbf{G}}^{\textbf{c}_{\textbf{x}_b}} \Vert \textbf{G}_{\textbf{x}_b} \textbf{c}_{\textbf{x}_b} -\textbf{h}_{\textbf{x}_b} \Vert_2^2   \nonumber \\
+\sum_t (\widetilde{\textbf{A}}_{coll} \textbf{c}_{\textbf{x}_b}+\textbf{s}_{coll}-\textbf{b}_{coll})^T \boldsymbol{\lambda}_{coll} \nonumber \\ +\rho_{coll}\Vert \widetilde{\textbf{A}}_{coll} \textbf{c}_{\textbf{x}_b}+\textbf{s}_{coll}-\textbf{b}_{coll}\Vert_2^2 \nonumber \\
+\sum_t {v}_{\phi} \lambda_{v_\phi}+ {w}_{\phi} \lambda_{w_\phi}+\rho_{v_\phi, w_\phi}\Vert \begin{bmatrix}
\cos(\phi)\\
\sin(\phi)\\
\end{bmatrix}- \begin{bmatrix}
{v}_{\phi}\\
{w}_{\phi}
\end{bmatrix}\Vert_2^2+ \phi_{b}\lambda_{\phi} \label{step_3_reform}
\end{align}
\normalsize

\noindent The solution steps for (\ref{step_2_reform})-(\ref{step_3_reform}) are summarized in Algorithm \ref{algo1} and \ref{algo2} respectively. Each optimization from (\ref{xi_update})-(\ref{phi_update}) involves minimizing a convex quadratic function ($l_2$ norm of an affine function+affine term). To reiterate, the Lagrange multipliers and quadratic penalties can be updated based on constraint residuals \cite{boyd_admm}. The non-negative slack variables $\textbf{s}_{coll}$ can be updated following the process presented in \cite{admm_qp}. Due to lack of space, we do not present the exact derivation.

\subsection{Notes on Computational Structure of Algorithm \ref{algo1} and \ref{algo2}} \label{notes}

\noindent \textbf{Distributiveness:} Consider optimization (\ref{xi_update}). It involves computing $\textbf{f}_m$  for ${^i}\textbf{x}_{e}$ at different instants of time and then minimizing the squared sum of all these functions. Now, importantly, ${^i}\textbf{x}_{e}$ at different time instants are independent of each other and thus, (\ref{xi_update}) can be split into $q$ parallel optimizations, where $q$ is the length of the time horizon. Similar parallel splitting can also be achieved for (\ref{vw_theta_update}). 


\noindent \textbf{Comparisons with Prior Work \cite{aks_iros18} :} 
Although the reformulations presented in our prior work \cite{aks_iros18} and the current proposed work are fundamentally different, certain comparisons can  still be drawn out. As shown earlier, the constraint functions $\textbf{f}_m$ are affine in $\textbf{v}_{\theta}, \textbf{w}_{\theta}$ when all other variables are held fixed. In contrast, in \cite{aks_iros18}, the non-linear constraints were affine in $v_{\theta_i}, w_{\theta_i} $, when all the other remaining elements in $\textbf{v}_{\theta}, \textbf{w}_{\theta}$ were held fixed. Thus, the proposed formulation induces multi affine structure over a larger set of variables. Furthermore, in \cite{aks_iros18}, approximation of squared acceleration cost was used to achieve the distributive structure.  In contrast, the current formulation achieves distributiveness by some clever use of trajectory parametrization in optimization (\ref{cost_task})-(\ref{collavoid}). Finally, \cite{aks_iros18} uses way-point parametrization for trajectories and thus, does not ensure higher order differentiability in the trajectories.

\small
\begin{algorithm*}
 \caption{Solving (\ref{step_2}) as a sequence of unconstrained QPs }\label{algo1}
    \begin{algorithmic}[1]
    
\State 
\small
\begin{eqnarray}
({^i}\textbf{x}_{e})^{k+1} = \arg\min \sum_{t, m = 1,2} \textbf{f}_m^T\boldsymbol{\lambda}_m+\rho_m\Vert \textbf{f}_m\Vert_2^2, \textbf{f}_1 = \textbf{P}(\textbf{c}_{\textbf{x}_b})^{k}+{\textbf{G}}_{\textbf{f}_1}({{^i}\textbf{x}_{e} }) \begin{bmatrix}
(v_{\phi})^k\\
(w_{\phi})^k\\
\end{bmatrix} -{^g}\textbf{x}_{d}, \textbf{f}_2 = \textbf{G}_{\textbf{f}_2}({{^i}\textbf{x}_{e}}) \begin{bmatrix}
(\textbf{v}_{{\theta}})^k\\
(\textbf{w}_{{\theta}})^k
\end{bmatrix}-\textbf{h}_{\textbf{f}_2} \label{xi_update}
\end{eqnarray}
\vspace{-0.3cm}

\vspace{-0.3cm}
\begin{eqnarray}
(\textbf{v}_{\theta})^{k+1}, (\textbf{w}_{\theta})^{k+1} =  \arg\min \sum_{t, m = 2} \textbf{f}_m^T\boldsymbol{\lambda}_m+\rho_m\Vert \textbf{f}_m\Vert_2^2\nonumber +\sum_t \textbf{v}_{\theta}^T \boldsymbol{\lambda}_{\textbf{v}_\theta} + \textbf{w}_{{\theta}}^T \boldsymbol{\lambda}_{\textbf{w}_\theta}+\rho_{\textbf{v}_\theta, \textbf{w}_\theta}\Vert \begin{bmatrix}
\textbf{v}_{\boldsymbol{\theta}}\\
\textbf{w}_{\boldsymbol{\theta}}
\end{bmatrix}-\begin{bmatrix}
\cos(\boldsymbol{\theta})^k\\
\sin(\boldsymbol{\theta})^k\\
\end{bmatrix}\Vert_2^2, \nonumber \\
\textbf{f}_2 = \textbf{G}_{\textbf{f}_2}( ({{^i}\textbf{x}_{e}})^{k+1} ) \begin{bmatrix}
(\textbf{v}_{{\theta}})\\
(\textbf{w}_{{\theta}})
\end{bmatrix}-\textbf{h}_{\textbf{f}_2} \label{vw_theta_update}
\end{eqnarray}

\vspace{-0.3cm}
\begin{eqnarray}
(\boldsymbol{\theta})^{k+1} = \arg\min \sum_{t,i} ( (\theta_i)-\textbf{p}(\textbf{c}_{\theta_i})^{k+1}) {\lambda}_{\textbf{c}_{\theta_i}}+  \rho_{\textbf{c}_{\theta_i}} (\theta_i-\textbf{p}(\textbf{c}_{\theta_i})^{k+1})^2+ \sum_t \rho_{\textbf{v}_\theta, \textbf{w}_\theta}\Vert \begin{bmatrix}
\cos(\boldsymbol{\theta})\\
\sin(\boldsymbol{\theta})\\
\end{bmatrix}-\begin{bmatrix}
(\textbf{v}_{{\theta}})^{k+1}\\
(\textbf{w}_{{\theta}})^{k+1}
\end{bmatrix}\Vert_2^2 +\boldsymbol{\theta}^T\boldsymbol{\lambda}_{\theta} \nonumber \\
=\arg\min \sum_{t,i} ( (\theta_i)-\textbf{p}(\textbf{c}_{\theta_i})^{k+1}) {\lambda}_{\textbf{c}_{\theta_i}}+  \rho_{\textbf{c}_{\theta_i}} (\theta_i-\textbf{p} (\textbf{c}_{\theta_i})^{k+1} )^2+ \sum_t \rho_{\textbf{v}_\theta, \textbf{w}_\theta}\Vert \boldsymbol{\theta}- \arctan2\frac{(\textbf{v}_{\theta})^{k+1}}{(\textbf{w}_{\theta})^{k+1}}  \Vert_2^2 +\boldsymbol{\theta}^T\boldsymbol{\lambda}_{\theta}\label{theta_update}
\end{eqnarray}
\small
\normalsize          
\end{algorithmic}  
\end{algorithm*}
        
\normalsize

\small
\begin{algorithm*}
 \caption{Solving (\ref{step_3}) as a sequence of unconstrained QPs }\label{algo2}
    \begin{algorithmic}[1]
    \small
\State 
\small
\begin{eqnarray}
(\textbf{c}_{\textbf{x}_b})^{k+1} = \arg\min w_2 J_{base}(\textbf{c}_{\textbf{x}_b})+\arg\min \sum_{t, m = 1,3} \textbf{f}_m^T\boldsymbol{\lambda}_m+\rho_m\Vert \textbf{f}_m\Vert_2^2 +(\textbf{G}_{\textbf{x}_b} \textbf{c}_{\textbf{x}_b} -\textbf{h}_{\textbf{x}_b})^T\boldsymbol{\lambda}_{\textbf{G}}^{\textbf{c}_{\textbf{x}_b}}+\rho_{\textbf{G}}^{\textbf{c}_{\textbf{x}_b}} \Vert \textbf{G}_{\textbf{x}_b} \textbf{c}_{\textbf{x}_b} -\textbf{h}_{\textbf{x}_b} \Vert_2^2 \nonumber \\
+\sum_t( \widetilde{\textbf{A}}_{coll} \textbf{c}_{\textbf{x}_b}+\textbf{s}_{coll}-\textbf{b}_{coll})^T \boldsymbol{\lambda}_{coll} +\rho_{coll}\Vert \widetilde{\textbf{A}}_{coll} \textbf{c}_{\textbf{x}_b}+\textbf{s}_{coll}-\textbf{b}_{coll}\Vert_2^2, \nonumber \\
\textbf{f}_1 = \textbf{P}\textbf{c}_{\textbf{x}_b}+{\textbf{G}}_{\textbf{f}_1}( ({{^0}\textbf{x}_{e} })^{k+1} ) \begin{bmatrix}
(v_{\phi})^{k}\\
(w_{\phi})^{k}\\
\end{bmatrix} -{^g}\textbf{x}_{d}, \textbf{f}_3 = \textbf{G}_{\textbf{f}_3} ((\textbf{c}_{\textbf{x}_b})) \begin{bmatrix}
(v_{\phi})^{k}\\
(w_{\phi})^{k}\\
\end{bmatrix}\label{cx_update}
\end{eqnarray}

\vspace{-0.3cm}
\begin{eqnarray}
(v_{\phi})^{k+1}, (w_{\phi})^{k+1} = \arg\min \sum_{t, m = 1,3} \textbf{f}_m^T\boldsymbol{\lambda}_m+\rho_m\Vert \textbf{f}_m\Vert_2^2 + \sum_t v_{\phi}\lambda_{v_\phi}+ {w}_{\phi} \lambda_{w_\phi}+\rho_{v_\phi, w_\phi}\Vert \begin{bmatrix}
\cos(\phi_{b})^k\\
\sin(\phi_{b})^k\\
\end{bmatrix}-\begin{bmatrix}
{v}_{\phi}\\
{w}_{\phi}
\end{bmatrix}\Vert_2^2 \nonumber \\
\textbf{f}_1 = \textbf{P}(\textbf{c}_{\textbf{x}_b})^{k+1}+{\textbf{G}}_{\textbf{f}_1}( ({{^0}\textbf{x}_{e} })^{k+1} ) \begin{bmatrix}
(v_{\phi})\\
(w_{\phi})\\
\end{bmatrix} -{^g}\textbf{x}_{d}, \textbf{f}_3 = \textbf{G}_{\textbf{f}_3} ((\textbf{c}_{\textbf{x}_b})^{k+1}) \begin{bmatrix}
v_{\phi}\\
w_{\phi}\\
\end{bmatrix}\label{vw_phi_update}
\end{eqnarray}
\vspace{-0.3cm}
\begin{eqnarray}
(\phi_{b})^{k+1} = \arg\min \rho_{v_\phi, w_\phi}\Vert \begin{bmatrix}
({v}_{\phi})^{k+1}\\
({w}_{\phi})^{k+1}
\end{bmatrix}-\begin{bmatrix}
\cos(\phi_{b})\\
\sin(\phi_{b})\\
\end{bmatrix}\Vert_2^2 +\lambda_{\phi}\phi_{b} \Rightarrow  \rho_{v_\phi, w_\phi}\Vert\phi_{b}-\arctan2\frac{({v}_{\phi})^{k+1}}{({w}_{\phi})^{k+1}} \Vert_2^2+\lambda_{\phi}\phi_{b} \label{phi_update}
\end{eqnarray}

\normalsize          
        \end{algorithmic}  
        \end{algorithm*}
        
\normalsize

\begin{figure*}[!h]
  \centering
\subfigure[]{
    \includegraphics[width= 4.05cm, height=3.5cm] {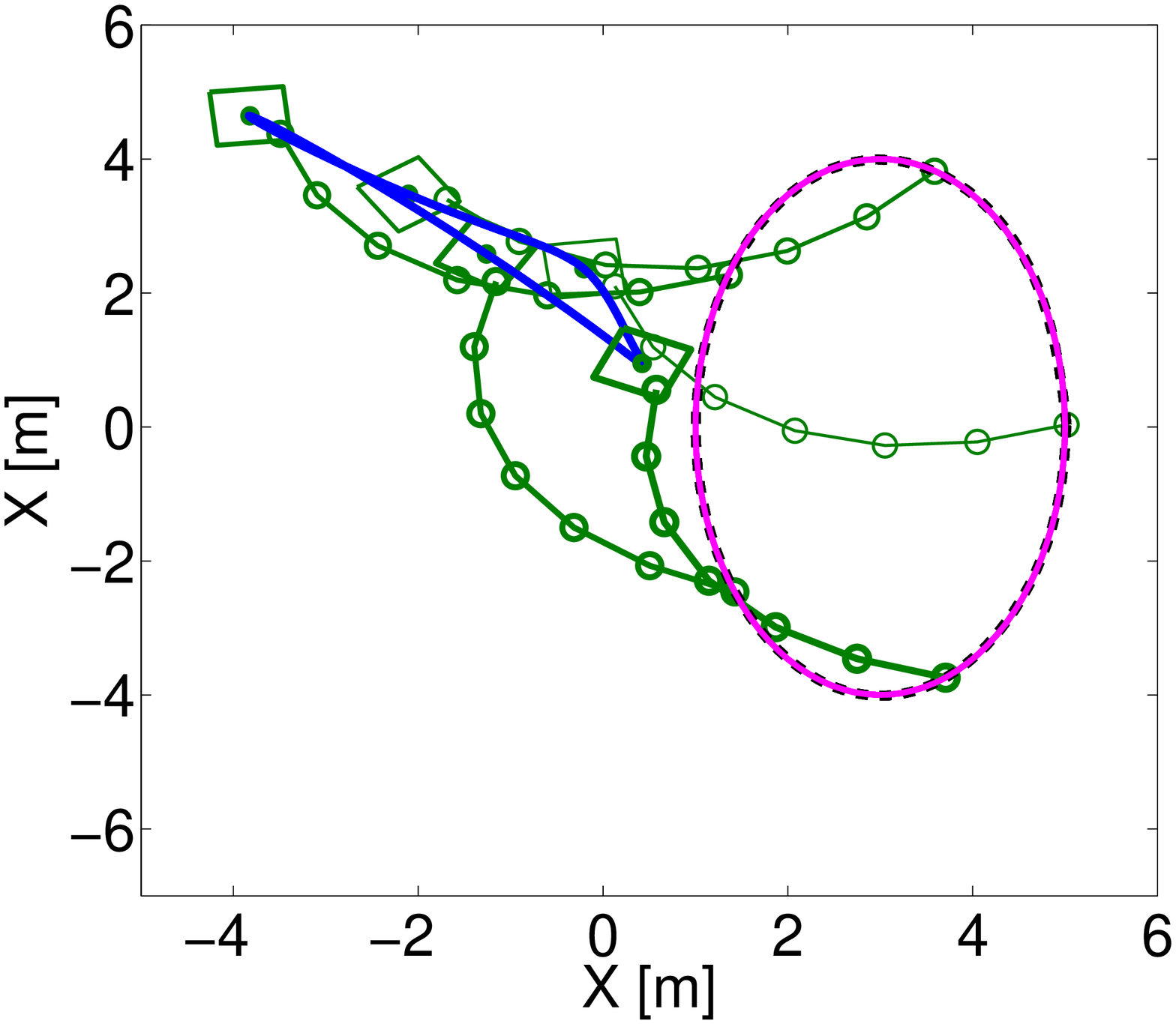}
    \label{planar_hol1}
   }\hspace{-0.60cm}   
\subfigure[]{
    \includegraphics[width= 4.05cm, height=3.5cm] {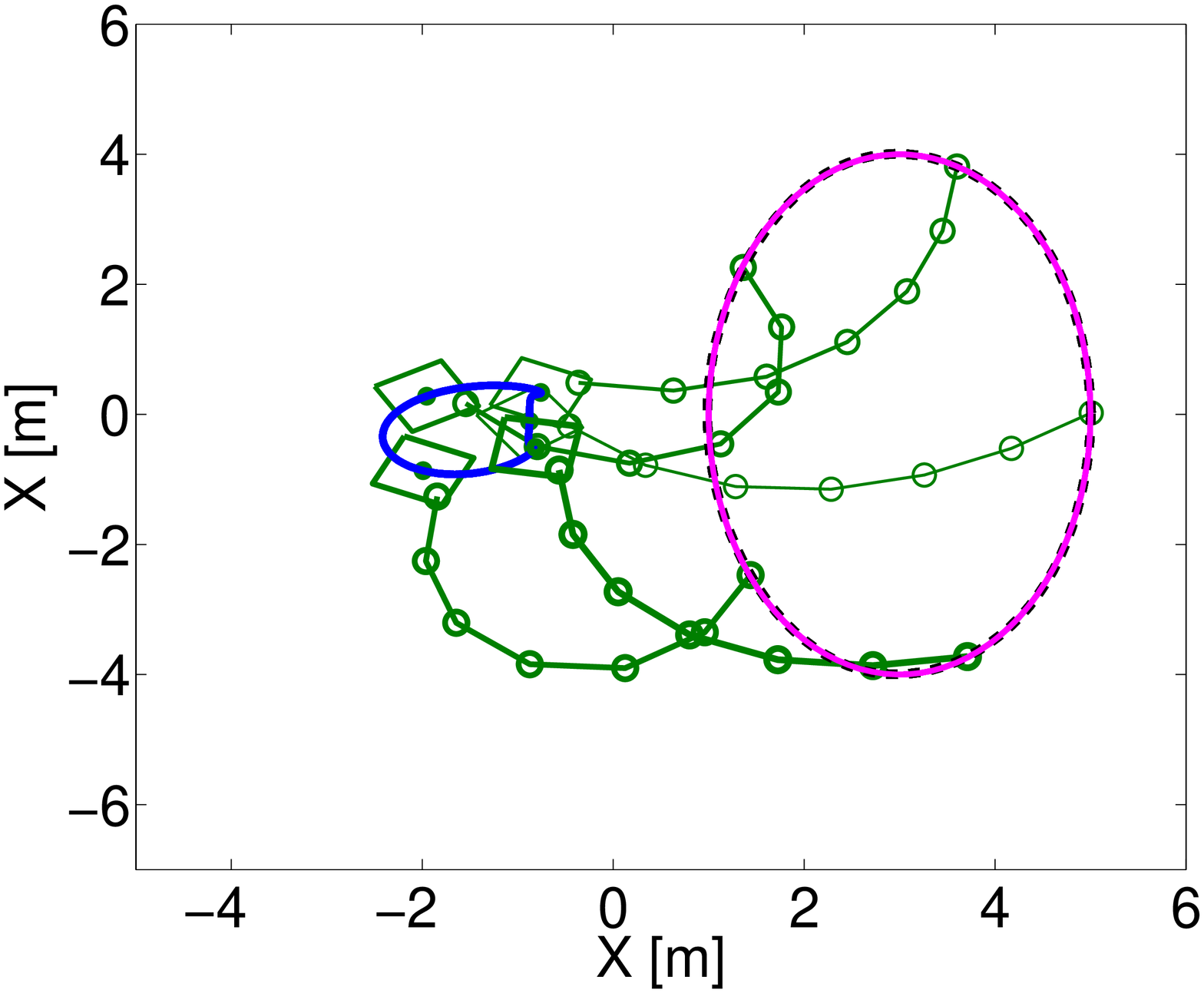}
    \label{planar_hol2}
   }\hspace{-0.70cm}
   \subfigure[]{
    \includegraphics[width= 6.50cm, height=3.5cm] {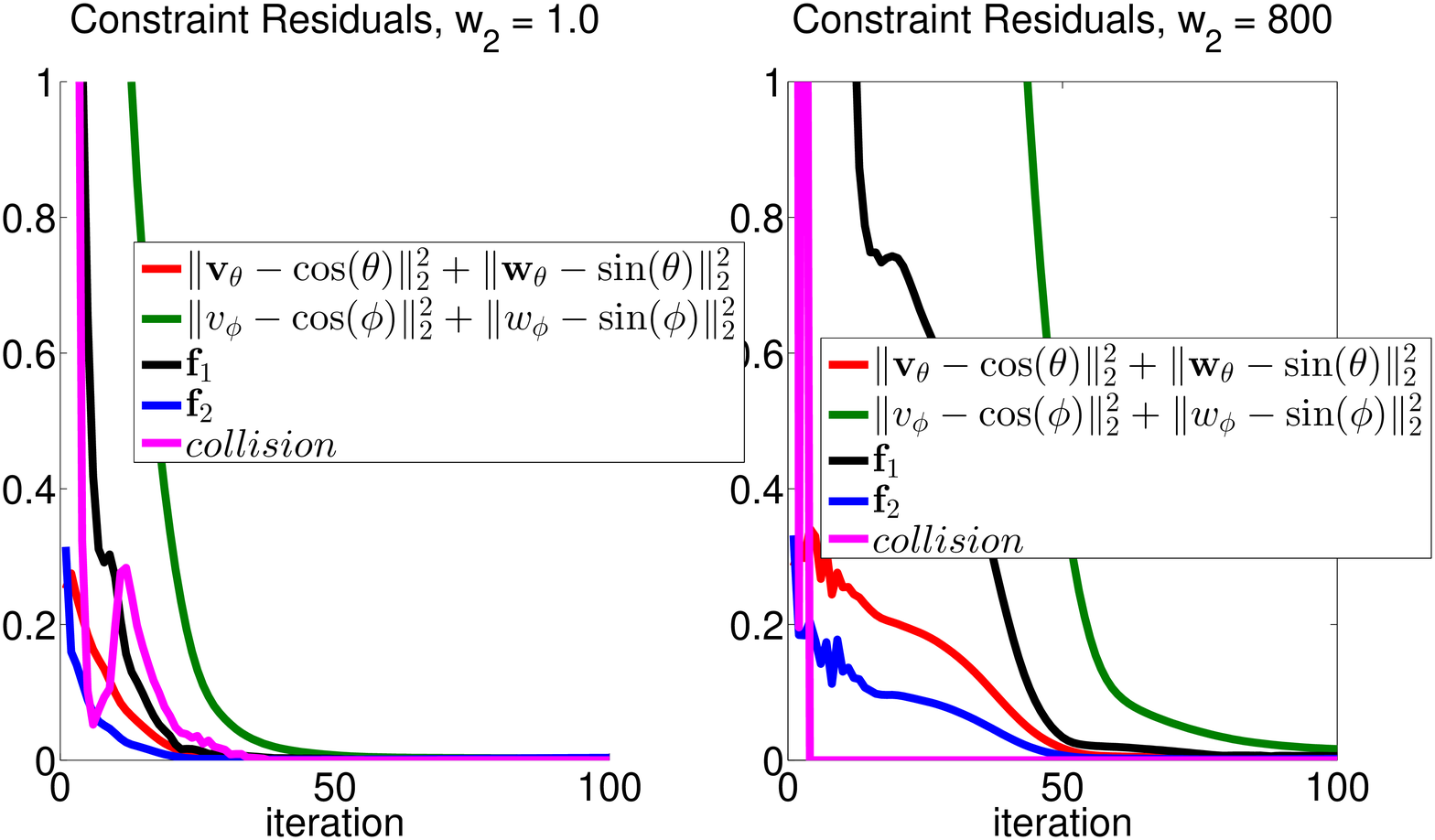}
    \label{constraint_residuals_planar_hol}
   }\hspace{-0.40cm}     
   \subfigure[]{
    \includegraphics[width= 3.50cm, height=3.5cm] {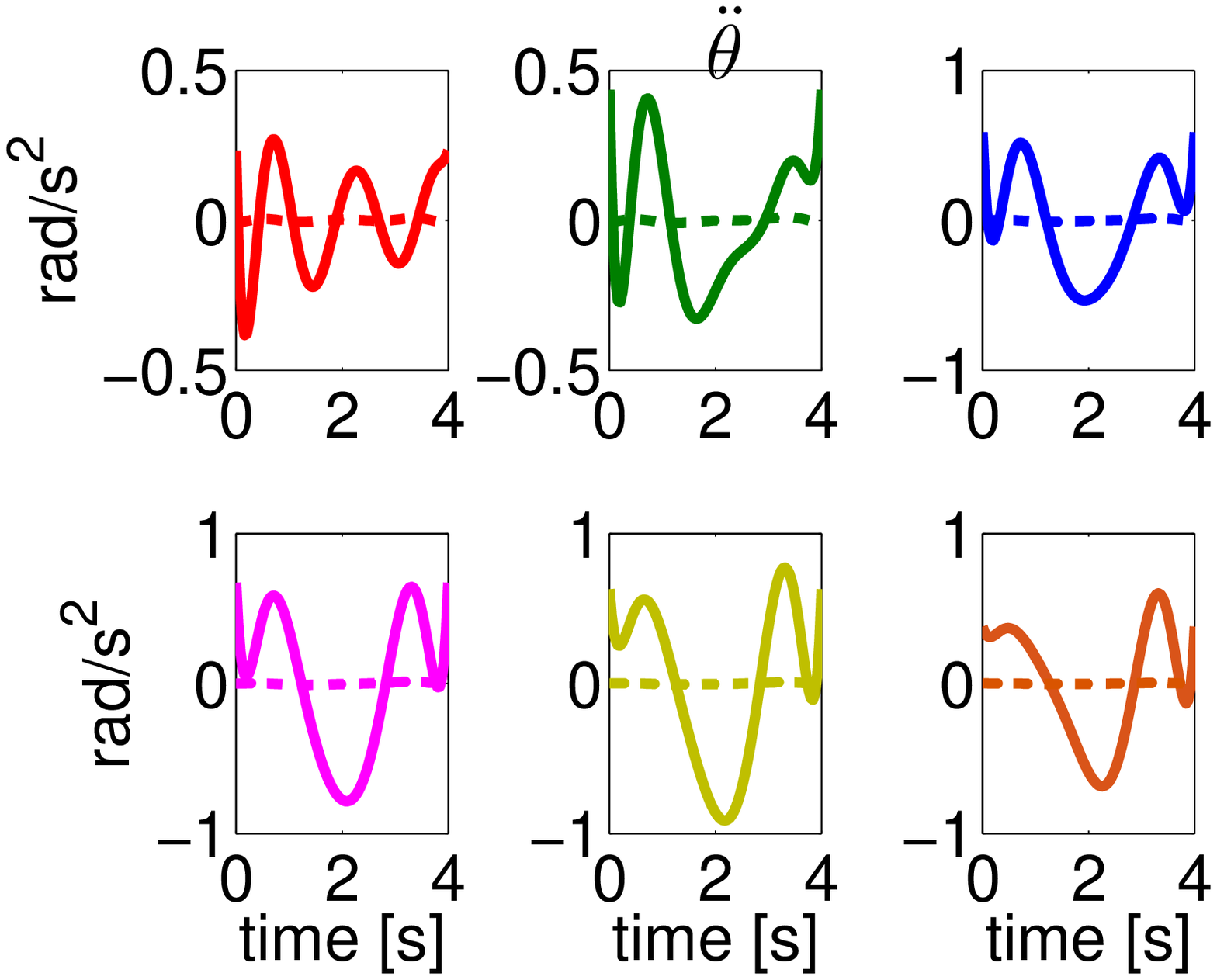}
    \label{theta_planar_hol}
   }   
   \vspace{-2ex}
\caption{(a) and (b): Simulations with a planar manipulator mounted on a holonomic base for $w_2=1$ and $w_2=800$ respectively (see (\ref{cost_task})). Desired end-effector trajectory (magenta) and the actual trajectory traced (black) are shown. Mobile base trajectory is shown in blue while the base and manipulator configurations at few time instants are shown in green. The darker traces correspond to configurations occurring later in time. The collision avoidance constraints required the mobile base to be outside the area bounded by the ellipse. (c): Relevant constraint residuals (refer text for details). (d) Manipulator joint acceleration for $w_2 = 1.0$ (dotted) and $w_2 = 800$ (solid).}
\vspace{-0.7cm}
\end{figure*}


\section{Simulation Results}
 \vspace{-1ex}
   
\begin{figure*}[!h]
  \centering
\subfigure[]{
    \includegraphics[width= 4.05cm, height=3.5cm] {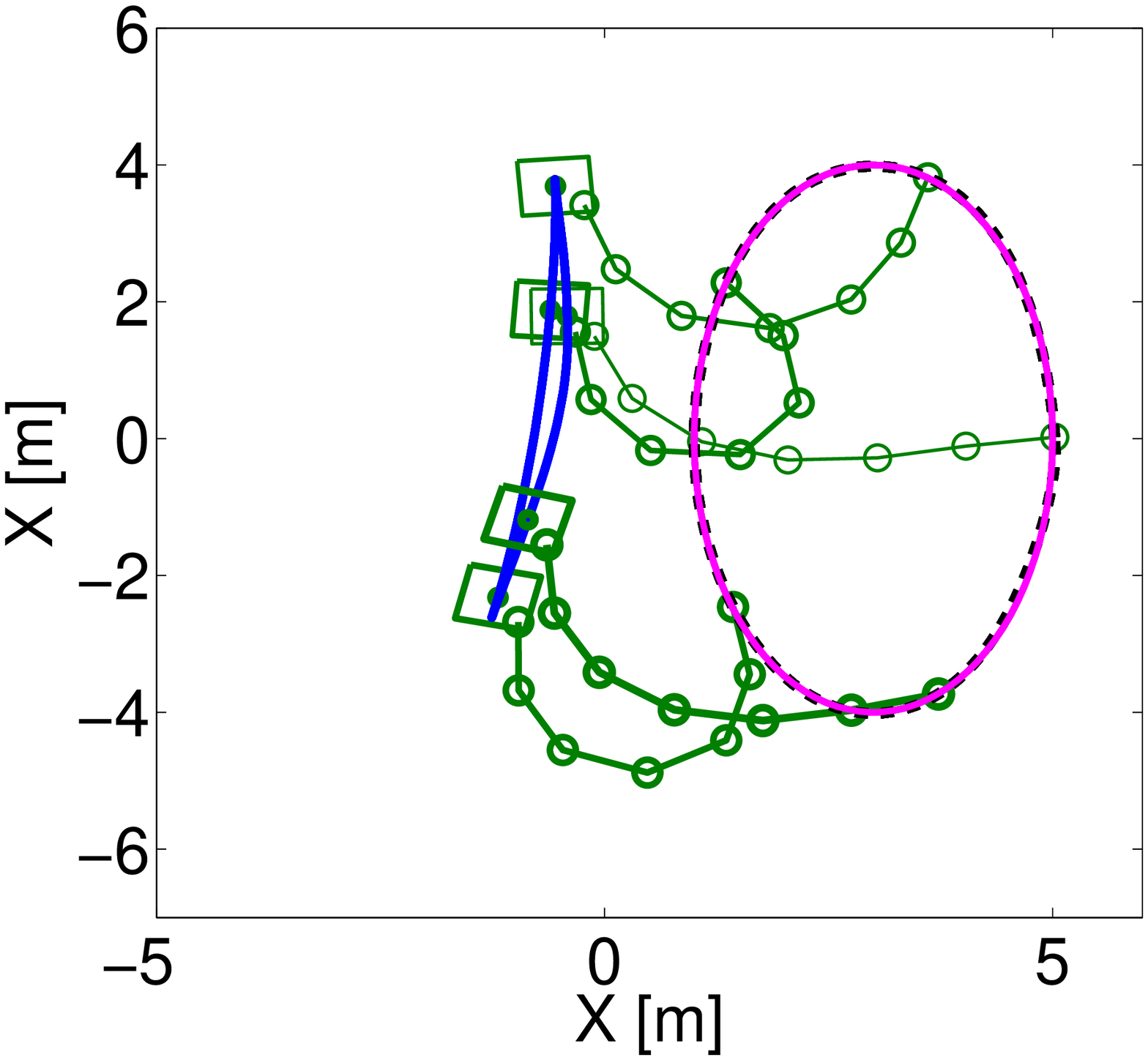}
    \label{planar_nonhol1}
   }\hspace{-0.60cm}   
\subfigure[]{
    \includegraphics[width= 4.05cm, height=3.5cm] {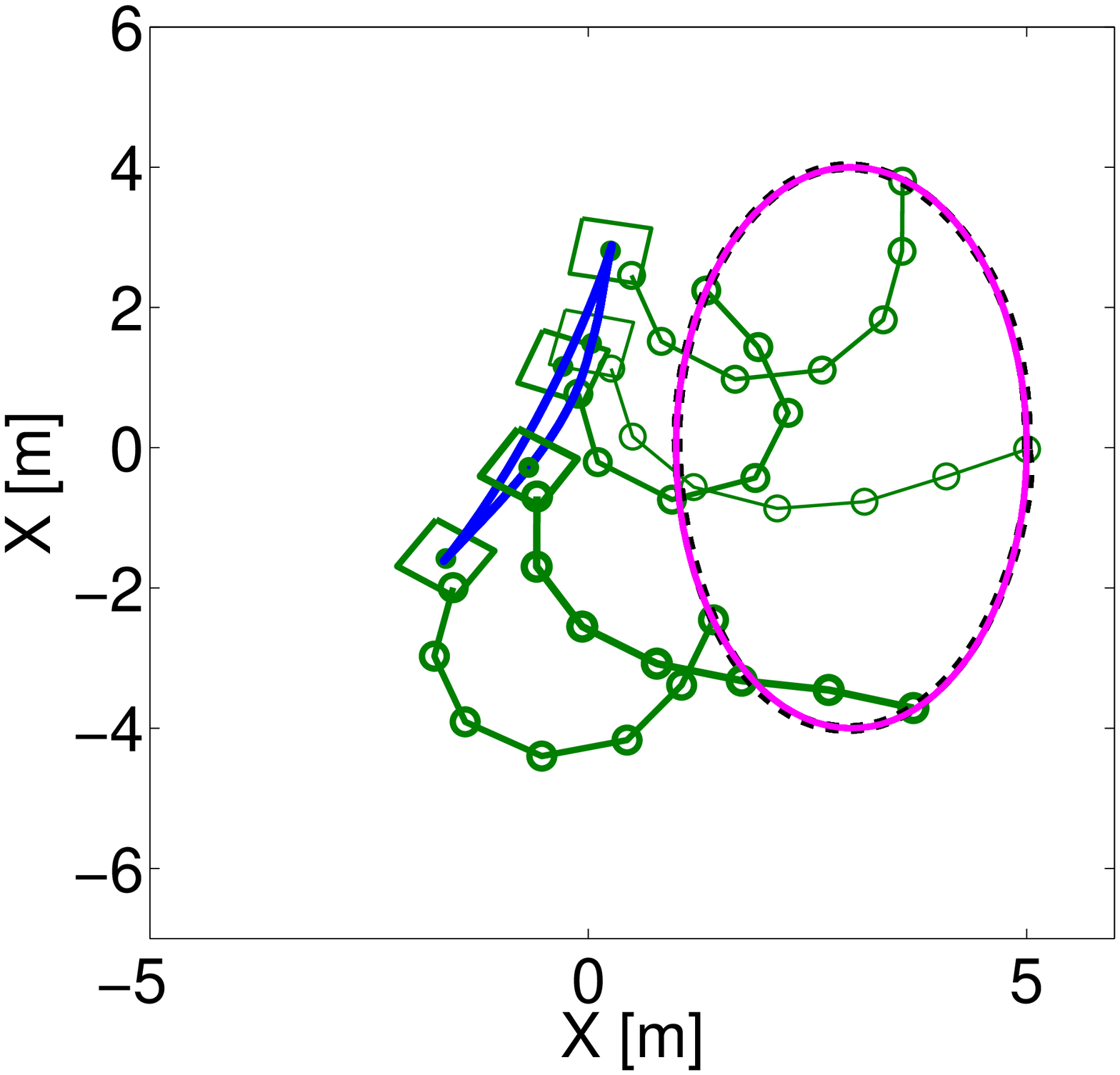}
    \label{planar_nonhol2}
   }\hspace{-0.70cm}   
   \subfigure[]{
    \includegraphics[width= 6.50cm, height=3.5cm] {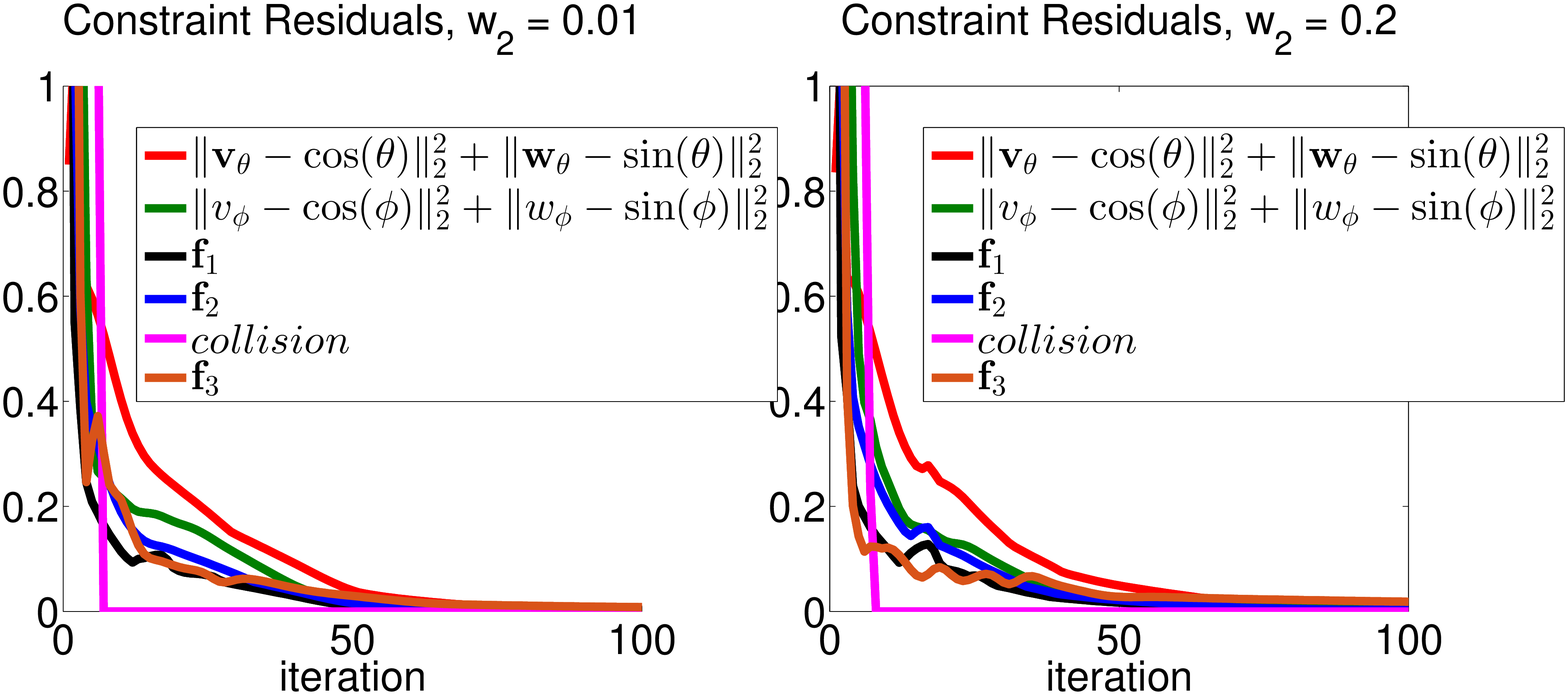}
    \label{constraint_residuals_planar_nonhol}
   }\hspace{-0.40cm}   
   \subfigure[]{
    \includegraphics[width= 3.50cm, height=3.5cm] {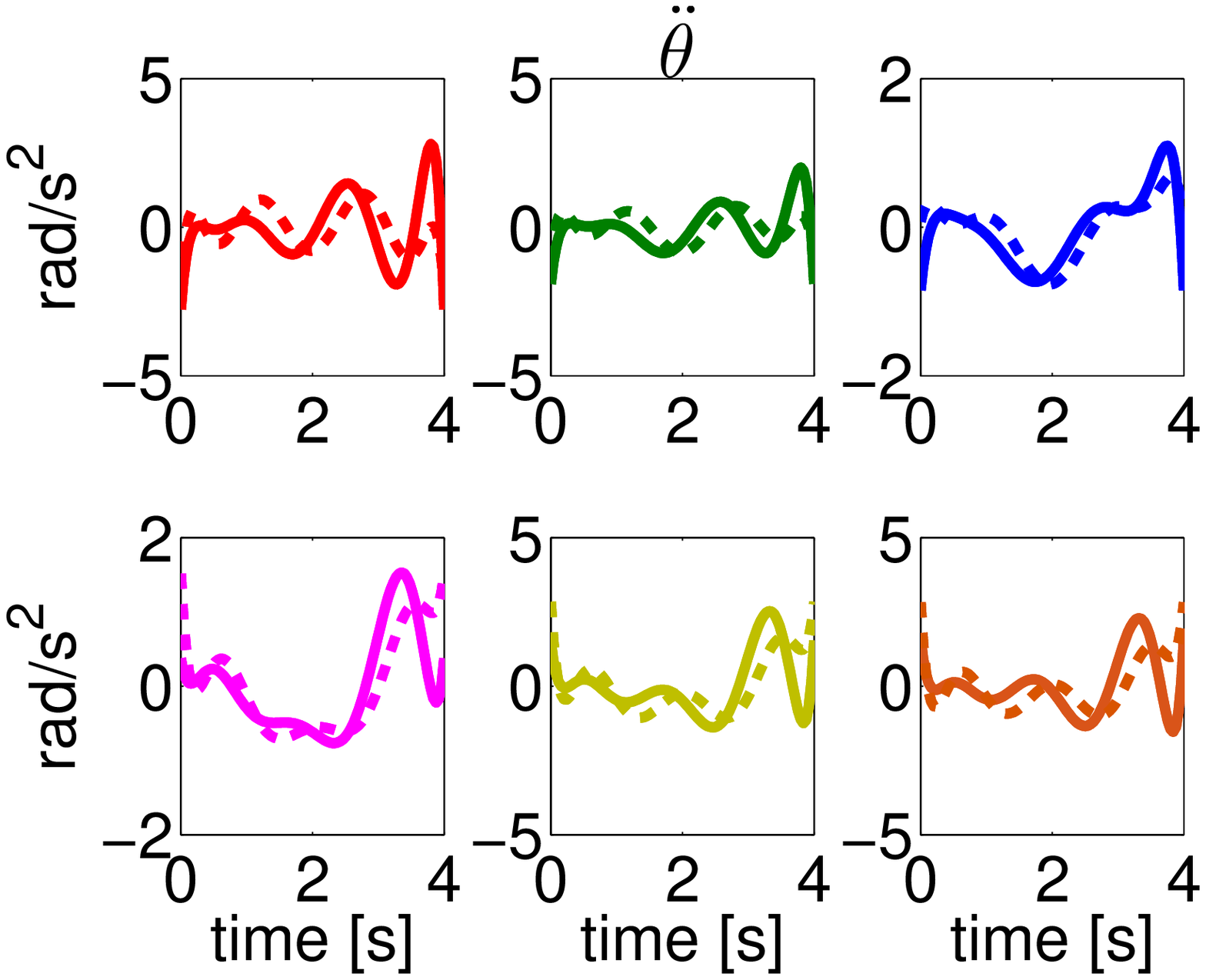}
    \label{theta_planar_nonhol}
   }\hspace{-0.50cm} 
\vspace{-0.5ex}
\caption{(a) and (b): Simulations with a planar manipulator mounted on a non-holonomic base for $w_2 = 0.01$ and $w_2 = 0.2$ respectively. The color notations are similar to Fig.\ref{planar_hol1}-\ref{planar_hol2}. (b) Relevant Constraint Residuals. Note that as compared to Fig.\ref{constraint_residuals_planar_hol}, here we have an additional constraint function $\textbf{f}_3$ modeling the no-lateral slip constraints. (d)Manipulator joint accelerations for $w_2 = 0.01$ (dotted) and $w_2 = 0.2$ (solid) }
\end{figure*}

\begin{figure*}[!h]
  \centering
\subfigure[]{
    \includegraphics[width= 6.05cm, height=3.05cm] {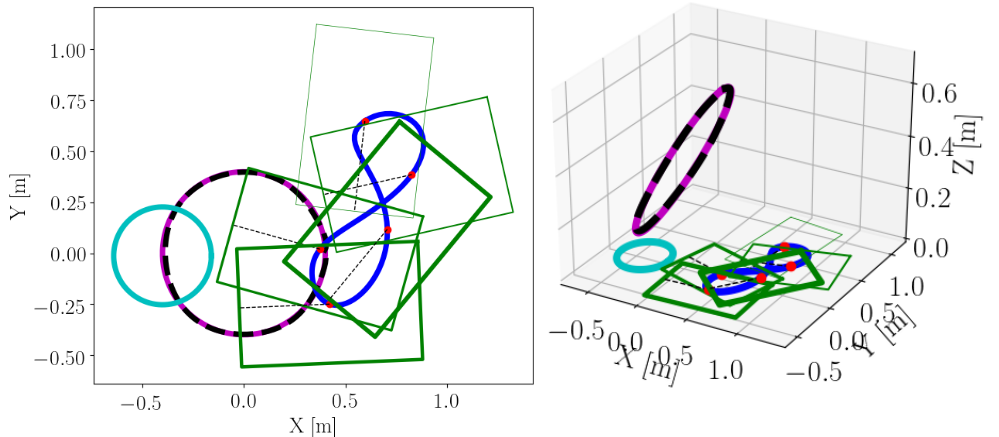}
    \label{franka_hol1}
   }\hspace{-0.30cm}   
\subfigure[]{
    \includegraphics[width= 3.05cm, height=3.05cm] {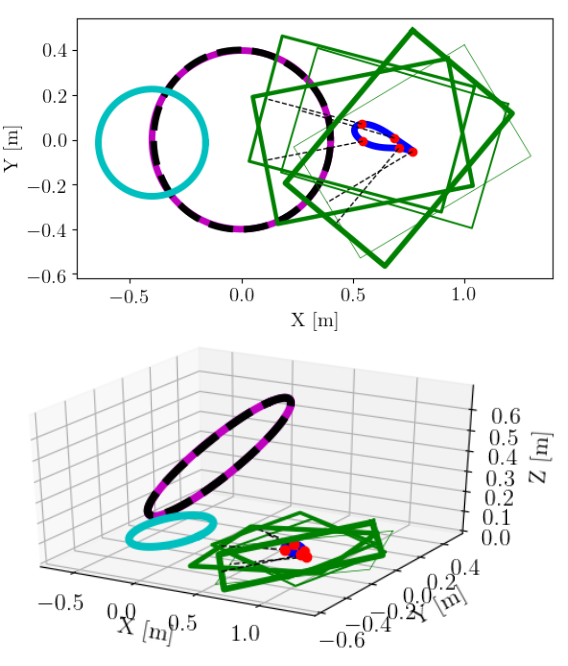}
    \label{franka_hol2}
   }\hspace{-0.50cm}   
\subfigure[]{
    \includegraphics[width= 5.00cm, height=3.05cm] {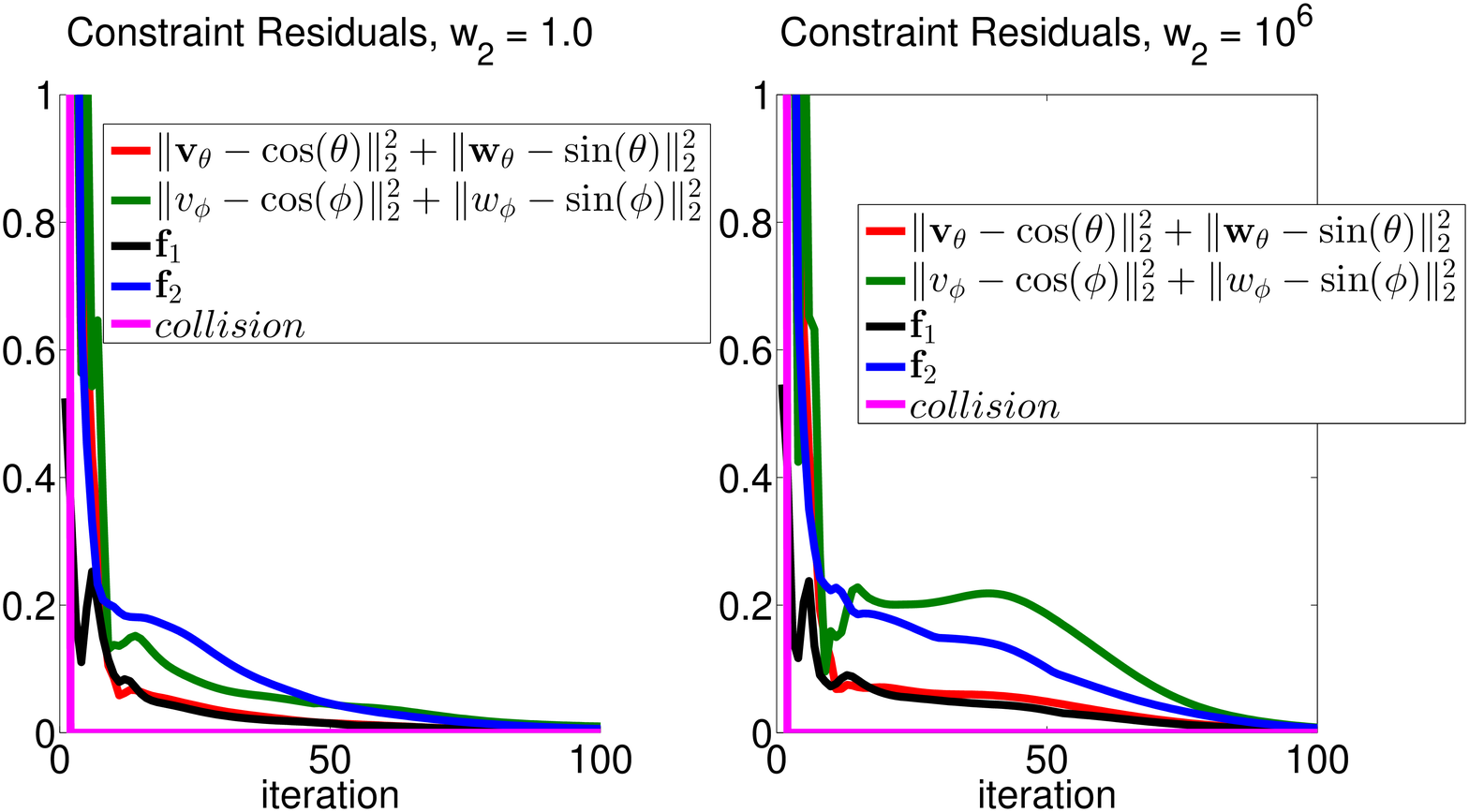}
    \label{constraint_residuals_franka_hol}
   }\hspace{-0.30cm}   
   \subfigure[]{
    \includegraphics[width= 3.05cm, height=3.05cm] {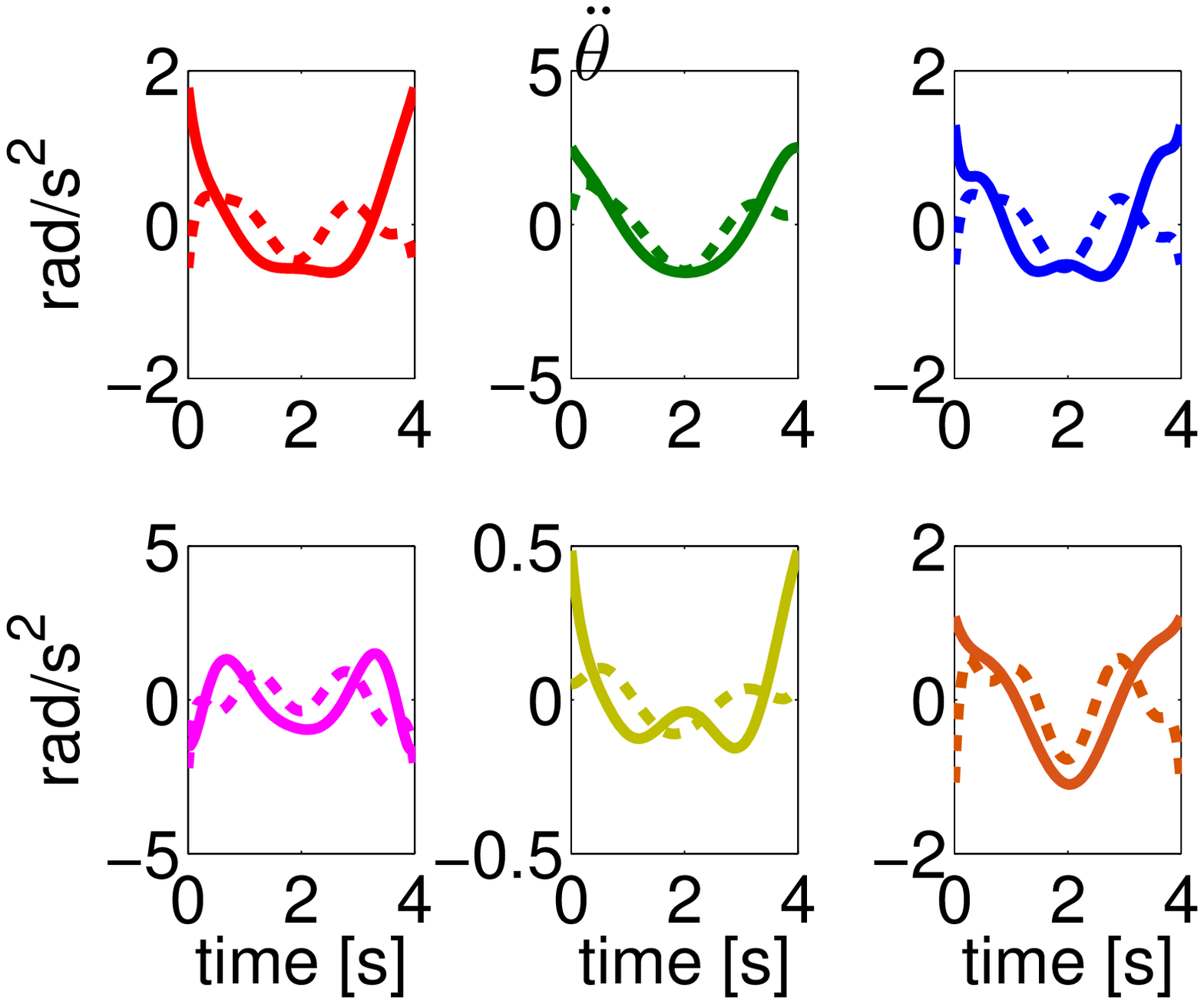}
    \label{theta_franka_hol}
   }
\caption{(a), (b): Simulations with a Franka Panda manipulator mounted on a holonomic base for $w_2 = 1.0$ and $w_2 = 10^6$ respectively. The color notations are same as the previous results. Additionally, the circle shown in cyan represents the obstacle in this case. (c) Relevant constraint residuals. (d) Manipulator joint acceleration profiles for $w_2 = 1.0$ (dotted) and $w_2 = 10^6$ (solid)  }   
\end{figure*}

\begin{figure*}[!h]
\centering
\subfigure[]{
    \includegraphics[width= 3.85cm, height=3.8cm] {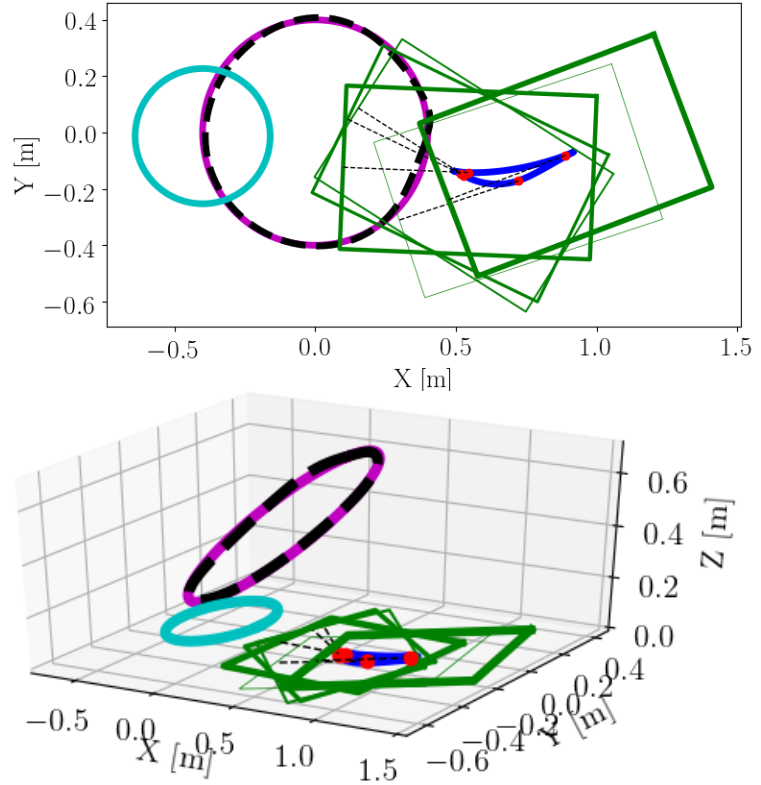}
    \label{franka_nonhol1}
   }\hspace{-0.60cm}
   \subfigure[]{
    \includegraphics[width= 3.85cm, height=3.8cm] {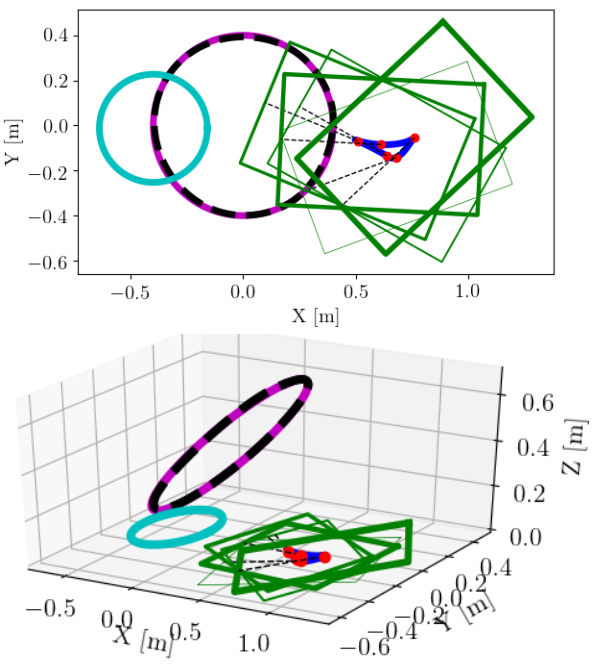}
    \label{franka_nonhol2}
   }\hspace{-0.60cm}
   \subfigure[]{
    \includegraphics[width= 6.20cm, height=3.8cm] {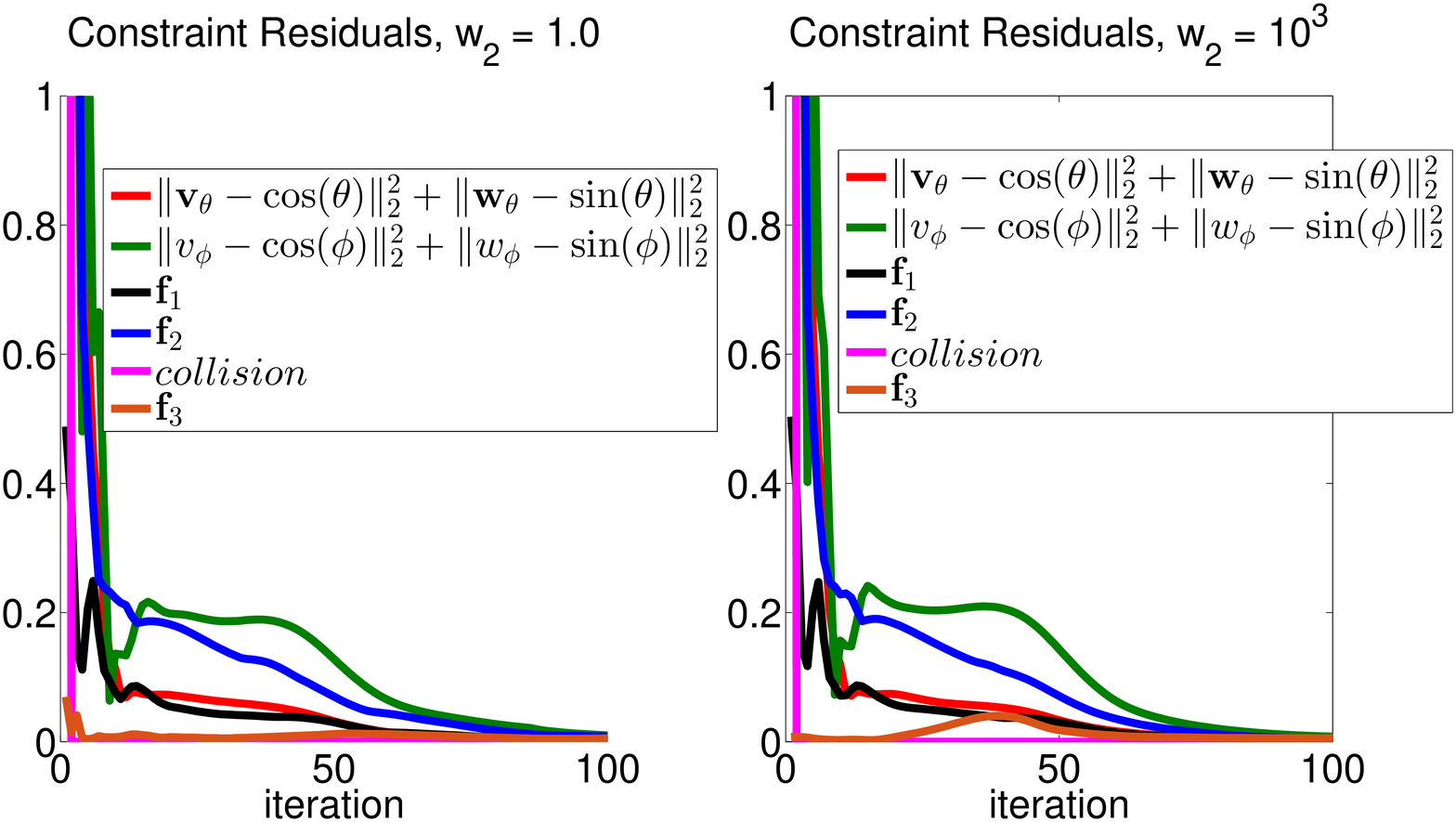}
    \label{constraint_residuals_franka_nonhol}
   }\hspace{-0.40cm}
   \subfigure[]{
    \includegraphics[width= 3.20cm, height=3.8cm] {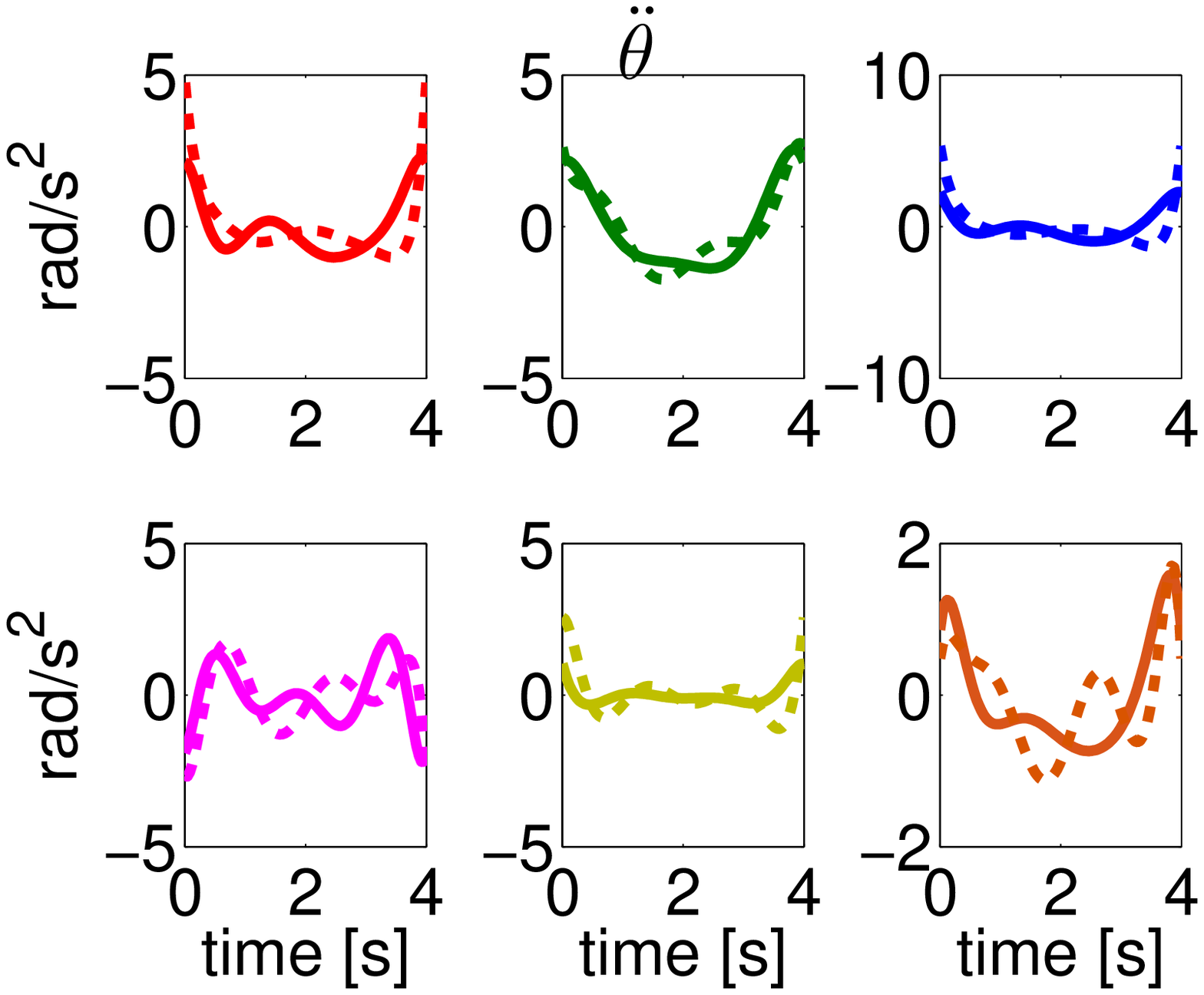}
    \label{theta_franka_nonhol}
   }\hspace{-0.60cm}
\caption{ (a), (b): Simulations with a Franka Panda manipulator mounted on a non-holonomic base for $w_2 = 1.0$ and $w_2 = 10^3$ respectively. The color notations are same as Fig. \ref{franka_hol1}-\ref{franka_hol2}. (c) Relevant constraint residuals. (d) Manipulator joint accelerations for $w_2 = 1.0$ (dotted) and $w_2 = 10^3$ (solid)  }
\end{figure*}

\begin{figure*}[!h]
\centering
\subfigure[]{
    \includegraphics[width= 6.5cm, height=5.05cm] {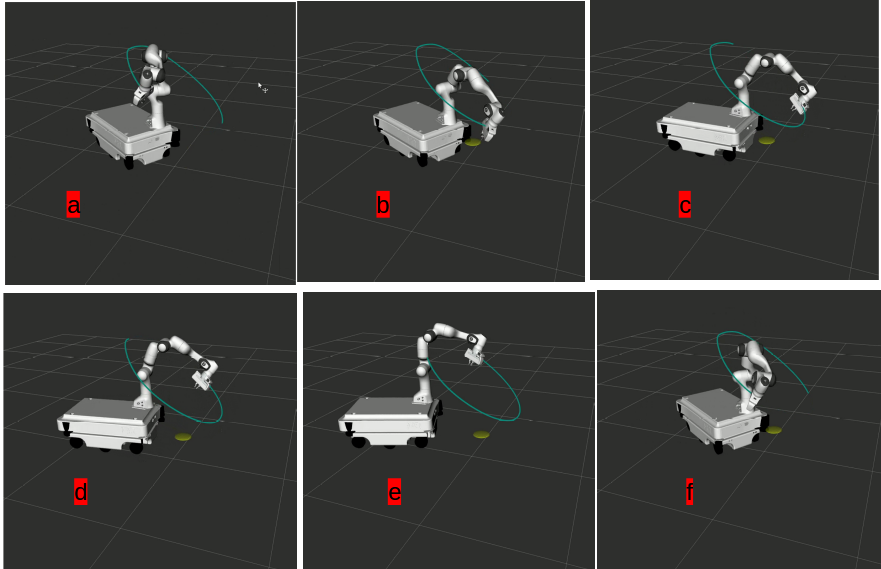}
    \label{snapshots}
   }\hspace{-0.80cm}
\subfigure[]{
    \includegraphics[width= 11.05cm, height=5.05cm] {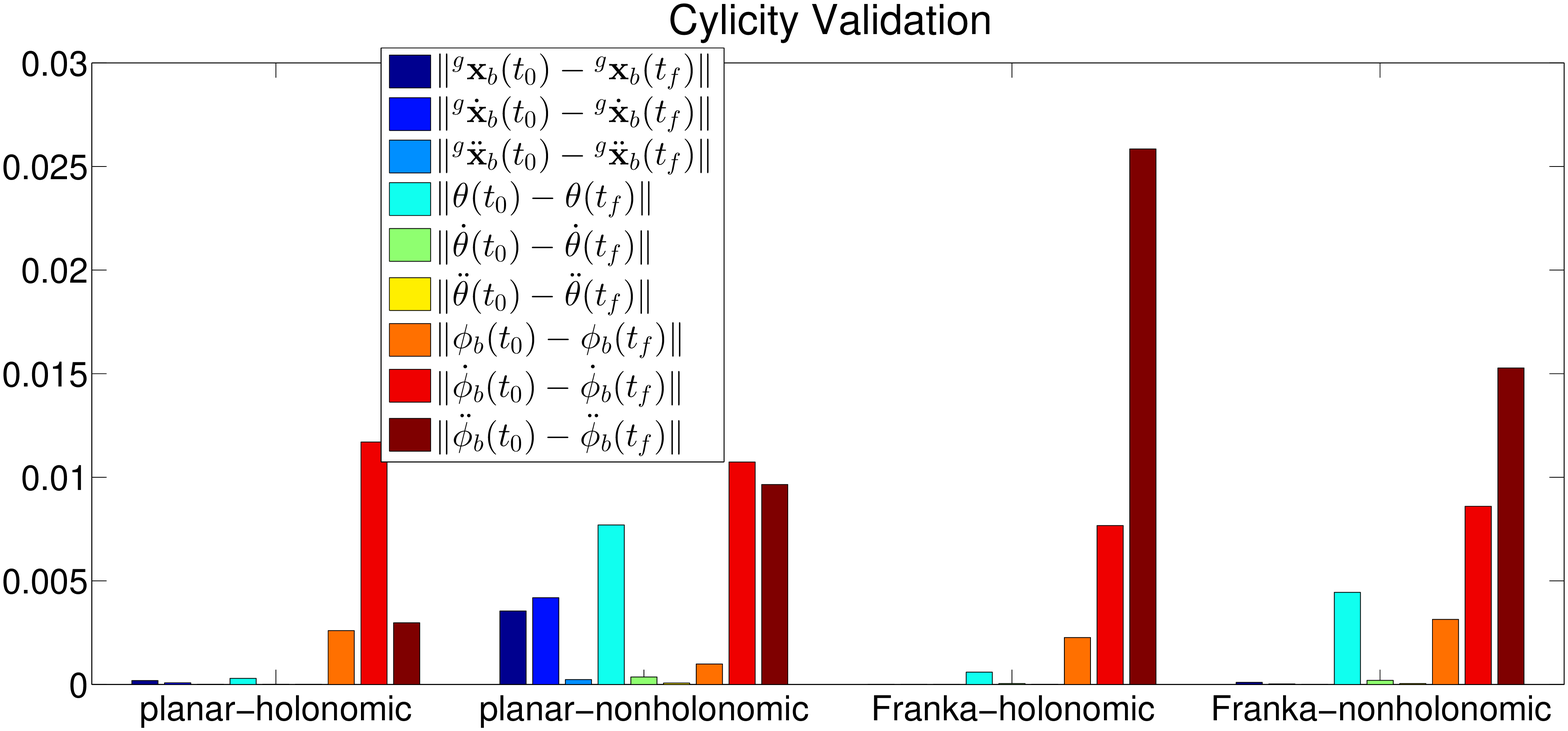}
    \label{cyclicity}
   }\hspace{-0.60cm}
\caption{(a): The Franka Panda arm mounted on a non-holonomic base executing the end-effector trajectory shown in Fig.\ref{franka_nonhol2}. For a cyclic end-effector trajectory, we obtain cyclic trajectories in the configuration space. Figure (b) validates this by showing the average residual between initial and final configurations, velocities, and accelerations for the trajectories shown in Fig. \ref{planar_hol1}-\ref{planar_hol2}, \ref{planar_nonhol1}-\ref{planar_nonhol2}, \ref{franka_hol1}-\ref{franka_hol2} and \ref{franka_nonhol1}-\ref{franka_nonhol2}.}
\vspace{-0.7cm}
\end{figure*}

\noindent \textbf{Set-up:} Simulations were performed on a $6$ degrees of freedom (dof) planar manipulator and 7 $dof$ Panda arm from Franka Emica (Fig.\ref{snapshots}). Motions for each of these manipulators were planned with a holonomic (Fig.\ref{planar_hol1}-(\ref{planar_hol2}), \ref{franka_hol1}-\ref{franka_hol2}) as well as a non-holonomic base (Fig. \ref{planar_nonhol1}-\ref{planar_nonhol2}, \ref{franka_nonhol1}-\ref{franka_nonhol2}). We only considered closed cyclic trajectories for the end-effector to highlight how our trajectory optimization overcomes the cyclicity bottleneck. For the planar manipulator case, the collision avoidance was modeled as the requirement that the mobile base be outside the area enclosed by the path traced by the end-effector (see Fig.\ref{planar_hol1}-\ref{planar_hol2}, \ref{planar_nonhol1}-\ref{planar_nonhol2}). For the simulations with Franka Panda arm, a circular obstacle region  (shown in cyan  in Fig.\ref{franka_hol1}-\ref{franka_hol2}, \ref{franka_nonhol1}-\ref{franka_nonhol2}) was considered. 

\noindent \textbf{Weight tuning:} We computed diverse set of trajectories by choosing $w_1 = 100$ and varying $w_2$ in (\ref{cost_task}). We gradually increased $w_2$ till we no-longer obtained feasible solutions for the specified iteration limit of 100. The results are summarized in Fig.\ref{planar_hol1}-\ref{planar_hol2}, \ref{planar_nonhol1}-\ref{planar_nonhol2}, \ref{franka_hol1}-\ref{franka_hol2}, \ref{franka_nonhol1}-\ref{franka_nonhol2}. Predictably, an increase in $w_2$ resulted in mobile base trajectories with shorter arc lengths. This in turn led to manipulator joint trajectories with higher acceleration magnitudes on average. (Fig.\ref{theta_planar_hol}, \ref{theta_planar_nonhol}, \ref{theta_franka_hol}, \ref{theta_franka_nonhol}). 


%

\noindent \textbf{Constraint Residuals: } Fig.\ref{constraint_residuals_planar_hol}, \ref{constraint_residuals_planar_nonhol}, \ref{constraint_residuals_franka_hol}, \ref{constraint_residuals_franka_nonhol} shows $\max({\Vert\textbf{f}_m}\Vert, \forall t)$ and maximum consensus residual (across all time) observed at each iteration. As shown, the residuals approach zero as the iterations progress, thus empirically verifying the convergence of our trajectory optimization. We observed that a higher $w_2$ adversely affected the rate of decrease of constraint residuals. This agrees with the trends observed in ADMM based approaches where optimality and feasibility compete with each other. We also observed that our trajectory optimization converged faster for holonomic base hinting at the complexity that stems from the differential non-holonomic constraints (\ref{nonhol_reform}).

\noindent \textbf{Smoothness:} Fig.\ref{theta_planar_hol}, \ref{theta_planar_nonhol}, \ref{theta_franka_hol}, \ref{theta_franka_nonhol} shows the acceleration profile of the manipulator joints. Clearly, these are smooth and differentiable. It is worth pointing out that existing trajectory optimizers like \cite{chomp}, \cite{trajopt} works with way-point parametrization and thus cannot ensure higher order smoothness. 

\noindent \textbf{Cyclicity:} We enforce cyclicity by defining appropriate boundary conditions to ensure that the initial and final configurations, velocities, and accelerations are same. Fig.\ref{cyclicity} shows the average of the relevant residuals for the holonomic and non-holonomic trajectories shown in Fig.\ref{planar_hol1}-\ref{franka_nonhol2}. The residuals for $\phi_b, \dot{\phi}_b$ are in the order of $10^{-2}$, while all others are in the order of $10^{-3}$. Low magnitudes of residuals validates that we indeed obtain (almost) cyclic trajectories in the configuration space for cyclic trajectories in the end-effector position space. It is worth pointing out that our trajectory optimization achieves low residuals in velocities and accelerations space as well. In contrast, works like \cite{kuka_cyclic_sample} focuses  only on the residual between the initial and final configurations. A slightly higher residual in $\dot{\phi}_b, \ddot{\phi}_b$ is due to the fact that we do not explicitly incorporate any boundary conditions on $\phi_b$ or its derivatives. Rather, we relied on the fact that $\phi_b$ is tightly coupled with the rest of the variables through the constraints on the end-effector path and the non-holonomic constraints (if applicable). This can be improved by parameterizing $\phi_b$ in the same way as $\theta_i$, ${^g}\textbf{x}_b$ in (\ref{smoothness_criteria_1}) and explicitly incorporating boundary conditions for it, although at the expense of slightly increasing the complexity of the trajectory optimization.

\begin{figure}[!h]
\centering
\subfigure[]{
    \includegraphics[width= 4.05cm, height=3.05cm] {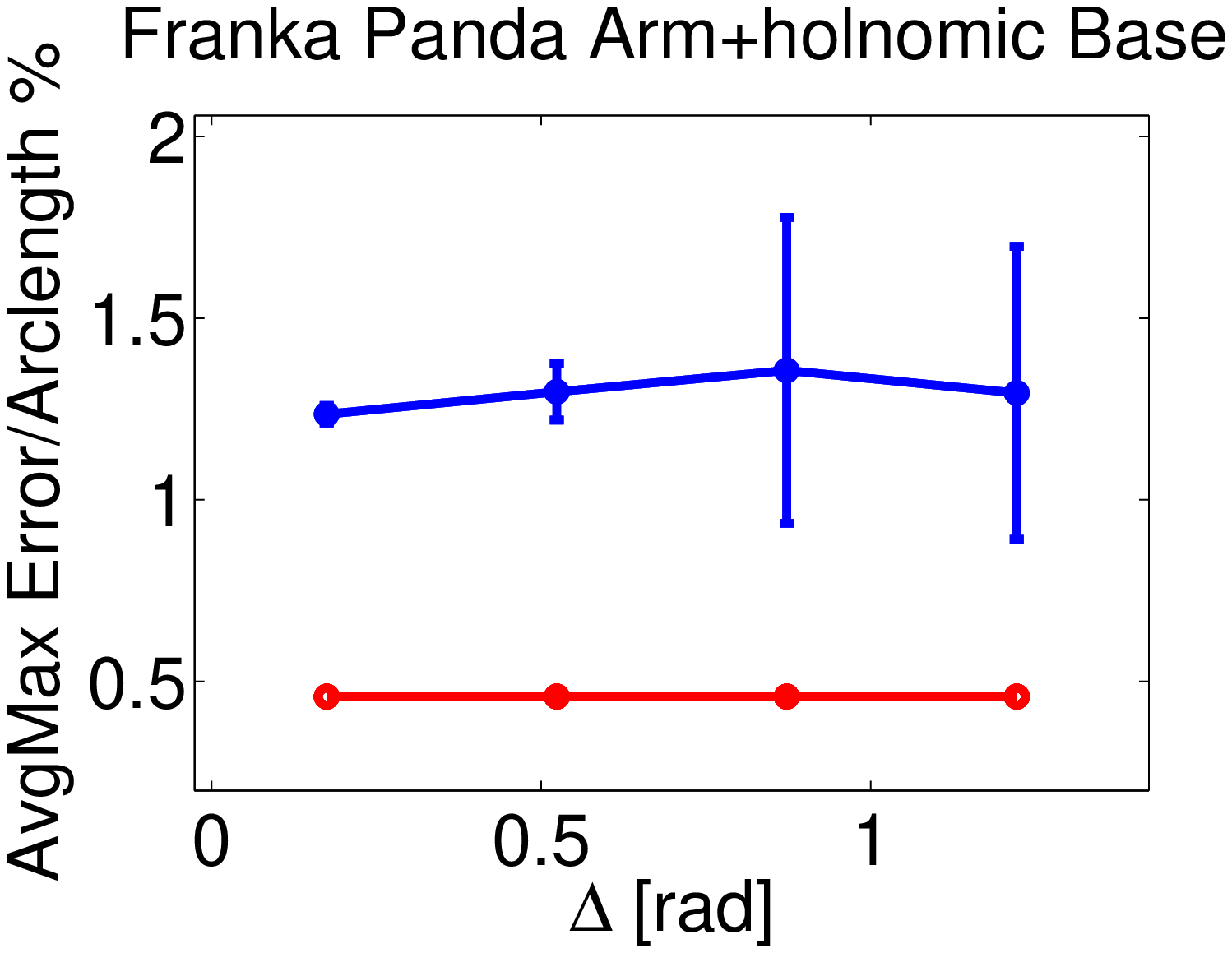}
    \label{perturb_franka_hol}
   }\hspace{-0.40cm}
\subfigure[]{
    \includegraphics[width= 4.05cm, height=3.05cm] {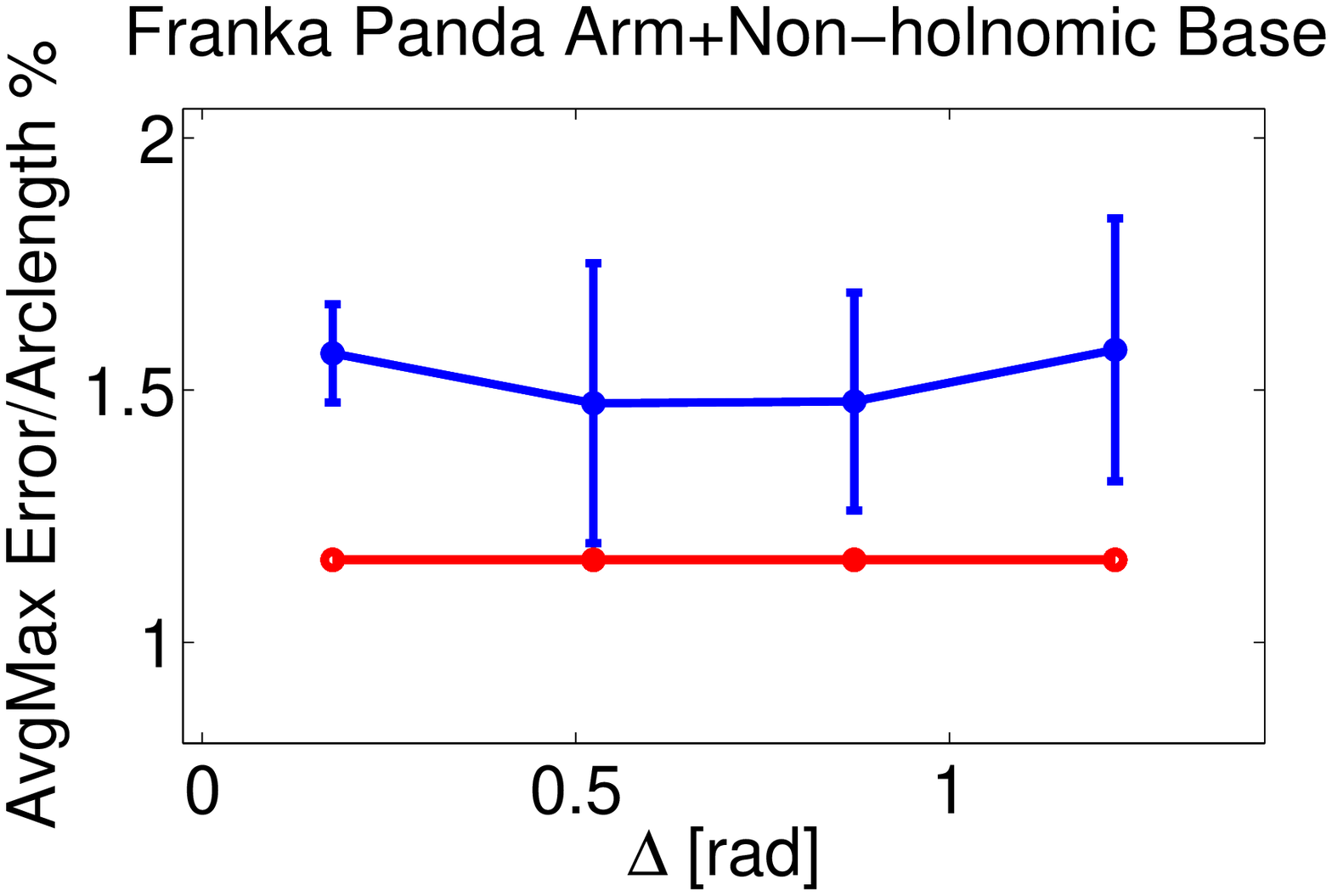}
    \label{perturb_franka_nonhol}
   }\vspace{-0.2cm}
\caption{Study of robustness of our trajectory optimization to poor initializations. The lines shown in blue represent the mean and standard deviation of normalized AvgMax error obtained with different perturbations $\Delta$ to the best initialization $\boldsymbol{\theta}_{init}$. The lines shown in red show the AvgMax error obtained for $\boldsymbol{\theta}_{init}$. Refer text for relevant definitions and explanations. }
\vspace{-0.5cm}
\end{figure}

\noindent \textbf{Robustness to Poor Initial Guess:} Non-convex optimizations often rely critically on the quality of the initial guess. To study the robustness of our trajectory optimization to poor initializations, we adopted the following approach. Let $\boldsymbol{\theta}_{init}$ be the initial guess for which the average of maximum end-effector position (AvgMax) error observed across a set of problem instances is the least. Let $\varepsilon$ represent a perturbation to $\boldsymbol{\theta}_{init}$ drawn from a uniform distribution $[-\Delta, \Delta ]$. We use the change in  AvgMax error with increase in $\Delta$ as our metric for robustness to poor initializations. The results obtained across $20$ problem instances (obtained by generating random ${^g}\textbf{x}_d$) involving a Franka Panda arm mounted separately on a holonomic and a non-holonomic base are presented in Fig.\ref{perturb_franka_hol} and \ref{perturb_franka_nonhol} respectively. The lines shown in blue represent the mean and standard deviation of AvgMax error for different perturbations, while the line shown in red presents the least AvgMax error obtained for $\boldsymbol{\theta}_{init}$. We normalized the errors by the arc length of ${^g}\textbf{x}_d$ (see Table \ref{sym_not}) and express it in percentage. Fig.\ref{perturb_franka_hol} shows that in the worst case, our trajectory optimization converges to a normalized AvgMax error of only $1.7 \%$, and $1.84 \%$ in Fig.\ref{perturb_franka_nonhol}.

\noindent \textbf{Computational Aspects:} The trajectory optimization was implemented in Python using Numpy libraries on a laptop with 12GB RAM, $i7$ processor with $2.5Ghz$ clock speed. For the planar manipulator case, we could compute a trajectory of 100 time steps in $4.0s$. For Franka manipulator, the same timing was around $5.0s$. Note that these timings were obtained without exploiting the distributive structure in the optimization. We believe that prototyping in a low level language like C++ coupled with parallelization can significantly improve the computation time.


\vspace{-0.3cm}

\section{Conclusions and Future Work}
In this paper, we presented a novel trajectory optimization for mobile manipulators with either holonomic or non-holonomic base. We successfully induced a multi-convex structure in this highly non-linear and non-convex problem. Our trajectory optimization solves the cyclicity bottleneck while achieving trajectories with any desired level of differentiability. Convergence was empirically validated by showing that constraint residuals go to zero as the optimization progress. Finally, robustness to poor initializations was also empirically verified.

There are several directions to expand the current work. Our preliminary evaluation shows us that each column of the manipulator Jacobian matrix can reformulated in a bi-affine form similar to (\ref{xn-1})-(\ref{x1}) and (\ref{bi_convex_fk}). Thus potentially, we can handle angular velocity constraints on the end-effector without disturbing the computational structure of our trajectory optimization. Posture constraints on the end-effector can be handled indirectly by constraining the angular velocities. We aim to explore this further in our future work.  Finally, a more formal understanding of the proposed convex surrogates along with their convergence analysis is also a key part of our future plans. 


\vspace{-0.2cm}
\bibliographystyle{IEEEtran}  
\bibliography{ref_iros19_short} 

\end{document}